\pdfoutput=1
\documentclass{scrartcl}
\usepackage[TS1, T1]{fontenc}
\usepackage{lmodern}
\usepackage[utf8]{inputenc}
\usepackage{amsmath}  %
\usepackage{tipa}  %

\usepackage{mathtools}
\usepackage{amssymb}
\usepackage{algpseudocode}
\usepackage{algorithm}
\usepackage{yfonts}
\usepackage{graphicx}
\usepackage{enumitem}
\usepackage{textcomp}
\usepackage{threeparttable}
\usepackage{subfigure}

\usepackage[
backend=biber,
style=alphabetic,
sorting=ynt
]{biblatex}

\newcommand*\longs{{\fontencoding{TS1}\selectfont s}}

\title{On the Accuracy of CRNNs for Line-Based OCR: A Multi-Parameter Evaluation}

\author{
  Bernhard Liebl \\
  \texttt{Computational Humanities Group, Leipzig University}\\
  \texttt{liebl@informatik.uni-leipzig.de}
  \and
  Manuel Burghardt \\
  \texttt{Computational Humanities Group, Leipzig University}\\
  \texttt{burghardt@informatik.uni-leipzig.de}
}
\date{\today}

\begin{document}

\maketitle

\begin{abstract}
We investigate how to train a high quality optical character recognition (OCR) model for difficult historical typefaces on degraded paper. Through extensive grid searches, we obtain a neural network architecture and a set of optimal data augmentation settings. We discuss the influence of factors such as binarization, input line height, network width, network depth, and other network training parameters such as dropout. Implementing these findings into a practical model, we are able to obtain a 0.44\% character error rate (CER) model from only 10,000 lines of training data, outperforming currently available pretrained models that were trained on more than 20 times the amount of data. We show ablations for all components of our training pipeline, which relies on the open source framework Calamari.
\end{abstract}

\section{Introduction}
Large parts of the humanities rely on printed materials as a source for their research. In order to leverage these printed sources for the digital humanities, they need to be represented in a machine-readable format, which is typically achieved by means of optical character recognition (OCR).
The quality of the OCR output is particularly important in the humanities, as they often form the basis for further analysis \cite{smith_ResearchAgendaHistorical_2018}.
This raises the question of how good OCR quality should be defined.
Reul et al. speak of "excellent character error rates (CERs) below 0.5\%" \cite{reul_OCR4allOpenSourceTool_2019}, which gives an indication of  typical distributions of CERs in many OCR contexts. More specific answers depend on various factors such as typefaces and paper quality. For modern typefaces, OCR is sometimes "considered to be a solved problem" \cite{reul_OCR4allOpenSourceTool_2019}. This is usually not the case for historical typefaces such as blackletter, which often expose a "highly variant typography" \cite{reul_OCR4allOpenSourceTool_2019} that renders OCR more difficult. In one evaluation by Reul et al. on 19th century blackletter prints, the best OCR model was able to achieve an overall mean CER of $0.61\%$\footnote{The CERs were between $0.01\%$ and $2.14\%$, depending on the material.} \cite{reul_StateArtOptical_2019}. On the other hand, a more recent study by Mart\'{i}nek et al. concludes that "taking into consideration a lower quality of historical scans and the old language, an acceptable value [for the CER] lies around 2\%".  Our own studies of 19th century blackletter text indicate that the CER of current state of the art models varies roughly between $0.8\%$ and $1.3\%$ (see our Section \ref{section:ablation-intro}). In this paper we want to investigate how these error rates from various studies can be systematically improved. %

One solution to improve the CER is post-processing, which is inherently difficult and error-prone. Smith and Cordell note that "[m]ost historical book and newspaper archives have deemed manual correction of texts too time consuming and costly" \cite{smith_ResearchAgendaHistorical_2018}. 
Another option is to train OCR models that produce good results right from the start.
However, generating the required amount of ground truth can be expensive and poses various problems, such as finding and following good transcription rules, human transcription errors, lack of tooling, etc. \cite{smith_ResearchAgendaHistorical_2018}. These problems tend to get harder with more ground truth.

In this paper, we investigate how we can leverage better network architectures and data augmentation to train better models with a minimum of training data, thereby alleviating the need to build huge collections of ground truth. In fact, we will show that a ground truth of 10,000 (10K) lines can produce models that give similar performance as models trained with 20K lines (CER $0.45\%$ vs. CER $0.42\%$).
Since our study incorporates approaches from various OCR architectures built for very different tasks - e.g. modern typeface OCR, historical typeface OCR, handwritten text recognition (HTR) and scene text detection -  we expect our findings to help bring substantial improvements to the quality of many OCR models in general. Although we focus on demonstrating that our proposed procedure excels in training high quality models for recognizing historical typefaces on highly degraded paper, all of the presented procedures should help enhance the accuracy of a wide variety of other OCR tasks, be it recognizing modern typefaces in books and magazines, or applying OCR to handwritten records.
The paper structure is twofold. In part 1, we show the influence of various factors such as neural network architectures and data augmentation using a small but representative corpus of about 3k lines, which allows us to run a large set of experiments. With the insights gained from the previous study, we then turn to a larger established corpus in part 2, in which we will demonstrate the effectiveness and reliability of our findings at a much larger scale.

\section{OCR in the Age of Deep Learning}
\label{section:Deep-OCR}
In existing research on OCR methods, one important distinction to be made is between older, character-based methods and more recent, line-based methods: the former segment single characters to recognize them one by one, the latter process whole text lines at once \cite{tong_MACRNNMultiscaleAttention_2020}. While character-based methods are still successfully used in scene text detection models \cite{chernyshova_TwoStepCNNFramework_2020}, most researchers agree that for recognizing printed text scanned in high resolutions line-based deep neural network (DNN) models are now the preferred state of the art solution \cite{breuel_HighPerformanceOCRPrinted_2013, wick_CalamariHighPerformanceTensorflowbased_2018, reul_StateArtOptical_2019, michael_D7HTREngine_2018, zhou_EASTEfficientAccurate_2017}. This widely employed architecture combines a Convolutional Neural Network (CNN) \cite{simonyan_VeryDeepConvolutional_2015} with some form of Recurrent Neural Network (RNN) - the latter is often a Long Short-Term Memory (LSTM) \cite{hochreiter_LongShortTermMemory_1997}, although Gated Recurrent Units (GRUs) have also been used successfully \cite{kiss_BrnoMobileOCR_2019}. For an overview of different LSTM architectures used in this context, such as Bidirectional LSTMs (BLSTMs) and Multidimensional LSTMs (MDLSTMs), see \cite{bukhari_InvestigativeAnalysisDifferent_2018}. The combination of CNN and RNN (e.g. LSTM or GRU) is usually referred to as Convolutional Recurrent Neural Network (CRNN) \cite{breuel_HighPerformanceText_2017}. This network is able to learn reading whole lines when trained via connectionist temporal classification (CTC) loss \cite{graves_ConnectionistTemporalClassification_2006}.  Various improvements to this basic architecture have been proposed. For example, Bluche et al. showed how to extend CTC with attention mechanisms to improve HTR performance \cite{bluche_ScanAttendRead_2017}. Another attention mechanism that extended BLSTMs was proposed by Tong et al.   \cite{tong_MACRNNMultiscaleAttention_2020}.  One recent advance in the overall quality of CRNN models is the introduction of ensemble voting by Wick et al. \cite{reul_ImprovingOCRAccuracy_2018}, which follows an established OCR method of combining multiple classifiers \cite{handley_ImprovingOCRAccuracy_1998}.

A considerable corpus of research exists on how to preprocess document scans through operations like despeckling, sharpening, blurring, or binarization, in order to improve the accuracy of existing OCR solutions; see Sporici et al. \cite{sporici_ImprovingAccuracyTesseract_2020} for a recent overview. A large part of this research is focused on providing good input for older, pre-CRNN and usually character-based OCR technology that often recognized single characters through computational geometry approaches and depended on some form of binarization \cite{smith_OverviewTesseractOCR_2007}.

Similar to Yousefi et al. \cite{yousefi_BinarizationfreeOCRHistorical_2015} we argue that the arrival of CRNNs has changed the situation considerably and the research focus should shift away from preprocessing images that are given to trained OCR models and instead turn to improving the CRNN model itself in order to obtain better results with a wide range of unprocessed inputs. One exception to preprocessing is deskewing which is still a very useful operation for CRNNs \cite{breuel_HighPerformanceOCRPrinted_2013}.

The objective of improving models implies training them in a better way. This involves making difficult decisions about factors that influence the OCR model's quality. Putting aside inference speed, and focusing fully on accuracy, we identify four central factors as research questions for this study\footnote{Note that at least one other crucial factor exists, which involves the overall training strategy and the employed optimizers \cite{martinek_TrainingStrategiesOCR_2019, reul_StateArtOptical_2019, michael_D7HTREngine_2018}. For the sake of simplicity, we use a standard procedure.}:

\begin{enumerate}
\item What kind of input data representation works best for training and inference, e.g should we binarize lines, and which line height should we use? See Section \ref{section:line-input-questions}.

\item What kind of network architecture is good? See Section \ref{section:crnn-arch}.

\item Which data augmentation works best? See Section \ref{section:data-augmentation-questions}.

\item How much training data - i.e. lines of annotated ground truth - is needed for training a model with good accuracy? See Section \ref{section:ablation}.
\end{enumerate}
The next section gives an overview of related for for each of the above research questions.

\section{Related Work}
\paragraph{Input data representation} The question of how to feed line images into a CRNN has seen some controversial research. For example, while some authors report that binarization improves OCR accuracy for line–based OCR (\cite{martinek_TrainingStrategiesOCR_2019}), others clearly argue against it \cite{michael_D7HTREngine_2018, yousefi_BinarizationfreeOCRHistorical_2015}\footnote{See Section \ref{section:binarization} for a more detailed discussion on this.}. The choice of line heights is another controversially discussed parameter: The classic MNIST dataset for training digit classifiers is 28 px high \cite{lecun_GradientBasedLearningApplied_1998}. In 2005, Jacobs et al. reported using a line height of 29 px "for performing OCR on low-resolution [webcam] images", using a CNN (but not CRNN) architecture \cite{jacobs_TextRecognitionLowresolution_2005}. Very recently, Chernyshova et al. built a CNN (but not CRNN) classifier, using a 19 px height for characters to detect "Cyrillic, Armenian, and Chinese text" in images captured by mobile phone cameras \cite{chernyshova_TwoStepCNNFramework_2020}. A study by Namsyl and Konya reports using a line height of 32 pixels that was determined experimentally. The authors report that "[l]arger sample heights did not improve recognition accuracy for skew-free text lines" \cite{namysl_EfficientLexiconFreeOCR_2019}. 
This 32 px limit is supported by Tong et al., who report that a height of 32 px produced the same accuracy as 48 px and 64 px, but that 16 px was too little to correctly recognize small characters \cite{tong_MACRNNMultiscaleAttention_2020}. On the other hand, Namsyl and Konya also state that "relatively long, free-form text lines [warrant] the use of taller samples" \cite{namysl_EfficientLexiconFreeOCR_2019}. This leads us to the situation with current state of the art (non-HTR) CRNN systems geared towards historical typefaces, where line heights of 40 px \cite{martinek_BuildingEfficientOCR_2020, mehrotra_CollaborativeDeepNeural_2019} and, more commonly, 48 px \cite{wick_CalamariHighPerformanceTensorflowbased_2018, breuel_HighPerformanceOCRPrinted_2013, kiss_BrnoMobileOCR_2019} are to be found\footnote{Since deskewing is a common practice for all CRNN systems \cite{breuel_HighPerformanceOCRPrinted_2013}, the line heights are rather high in terms of what Namsyl and Konya \cite{namysl_EfficientLexiconFreeOCR_2019} and Tong et al. \cite{tong_MACRNNMultiscaleAttention_2020} suggest.}. In the context of HTR, line heights between 60 px and 80 px have been used \cite{michael_D7HTREngine_2018, wigington_DataAugmentationRecognition_2017}. In general, HTR line heights tend to be larger than in OCR.

\paragraph{Network architectures} A large number of network architectures for OCR have been proposed. Calamari and its pretrained models by default employ two convolutional layers with different filter sizes and a single LSTM layer \cite{wick_CalamariHighPerformanceTensorflowbased_2018}. Tesseract's default training seems to employ one convolutional layer followed by multiple stacked LSTMs\footnote{See \url{https://github.com/tesseract-ocr/tesstrain} and \url{https://tesseract-ocr.github.io/tessdoc/VGSLSpecs}}. Both systems seem to use these settings for both modern and historical typefaces. In a recent study Mart\'{i}nek et al. proposed a network architecture for historical documents using two convolutional layers with 40 filters each and two stacked BLSTMs \cite{martinek_BuildingEfficientOCR_2020}. Transkribus HTR+ employs three convolutional layers and three stacked BLSTM layers \cite{michael_D7HTREngine_2018}, whereas an HTR system by Wigington et al. uses 6 convolutional layers and two BLSTM layers \cite{wigington_DataAugmentationRecognition_2017}. In the context of text detection in scenes, the EAST detector by Zhou et al. experimented with various established CNN architectures such as VGG-16 \cite{zhou_EASTEfficientAccurate_2017, simonyan_VeryDeepConvolutional_2015}, an approach that was also taken more recently by Ki\v{s}\v{s} et al. for the detection of text in photos taken by mobile devices \cite{kiss_BrnoMobileOCR_2019}. In a recent system aimed "to overcome the limitations of both printed- and scene text recognition systems", Namysl and Konya propose a fully convolutional network with 10 convolutional layers and one pooling layer \cite{namysl_EfficientLexiconFreeOCR_2019}. Similarly, Mehrotra et al. propose an architecture based on VGG-13, consisting of 10 convolutional layers and 5 pooling layers, for recognizing Indian languages \cite{mehrotra_CollaborativeDeepNeural_2019, simonyan_VeryDeepConvolutional_2015}. According to Breuel, OCR and HTR warrant slightly different network designs and goals \cite{breuel_HighPerformanceText_2017}. Unfortunately, there seems to be no single survey study comparing all these different models for line-level OCR under similar conditions and using similar typefaces. It is therefore somewhat unclear how effective one architecture is versus another one and why one set of hyperparameters (e.g. filter size, depth of network) was chosen over another. While methods for comparing different architectures and finding optimal ones in a large search space of possibly suitable candidates have been investigated in many forms in the larger DNN community\footnote{For example, several CNN architectures have been designed for (non-OCR) classifications tasks -- such as CIFAR-10 -- in this manner \cite{zoph_NeuralArchitectureSearch_2017}.} -  a task often referred to as Neural Architecture Search (NAS) \cite{elsken_NeuralArchitectureSearch_2019} - there seems to be little related research in terms of finding optimal CRNN architectures for line-based OCR.

\paragraph{Data augmentation} The existence of data augmentation for training CRNN models for OCR has often been acknowledged and described in some detail \cite{martinek_TrainingStrategiesOCR_2019, wick_CalamariHighPerformanceTensorflowbased_2018, michael_D7HTREngine_2018}. It is rarer though to find exact numbers that measure the impact of data augmentation on the overall accuracy. One example is Wigington et al., who describe two detailed data augmentations with specific parameters that "produce the lowest word error rate (WER) to date over hundreds of authors, multiple languages, and thousands of documents" in the context of HTR \cite{wigington_DataAugmentationRecognition_2017}. Most research that touches the topic however describes highly coalesced augmentation steps that combine a plethora of different methods, which makes it impossible to fathom the usefulness of or damage done by one data augmentation method versus another. For example,  Michael et al. give a detailed description on their data augmentation as consisting of "affine transformations, dilation and erosion, as well as elastic and grid-like distortions" \cite{michael_D7HTREngine_2018}. Similarly, Namysl and Konya use "Gaussian smoothing, perspective distortions, morphological filtering, downscaling, additive noise, and elastic distortions" as well as a new augmentation called "alpha compositing" \cite{namysl_EfficientLexiconFreeOCR_2019}. Martinek et al. create training data synthetically by concatenating letter symbols \cite{martinek_BuildingEfficientOCR_2020, martinek_TrainingStrategiesOCR_2019}, but it is not clear how these methods fare in comparison to or in combination with other data augmentation methods. While there has been detailed research on the effectiveness of data augmentation policies with regard to CNN performance - especially measuring single-character classification tasks (using MNIST) and non-OCR classification tasks (such as CIFAR-10) \cite{shorten_SurveyImageData_2019, taylor_ImprovingDeepLearning_2018, cubuk_AutoAugmentLearningAugmentation_2019, simard_BestPracticesConvolutional_2003, hu_PreliminaryStudyData_2019, lecun_GradientBasedLearningApplied_1998} - little research seems to be available in the context of CRNNs for complex OCR tasks, e.g. using historical typefaces with a large alphabet. 

\paragraph{Training set size} The fourth and last aspect to be discussed here is the amount of necessary training data. One state of the art benchmark for this question was recently provided by Ströbel et al., who obtained a CER of $0.48\%$ by training a model through 38,756 annotated lines from a historical newspaper \cite{strobel_HowMuchData_2020, strobel_ImprovingOCRBlack_2019}. Since the result was achieved by employing the closed source system Transkribus HTR+, it is difficult to gain deeper insights about the interplay of network architecture, training set size and employed data augmentation in this case. Our study will show that very similar results can be obtained with only about one fourth of the training data by using the open source system Calamari with a custom training pipeline (see Section \ref{section:ablation} for details). Note that Reul et al. showed that any CRNN training pipeline can reduce the amount of training data and improve quality by using transfer learning \cite{reul_TransferLearningOCRopus_2017}.

\bigskip

In summary, related work on OCR training procedures often collates various important aspects of OCR training, but only measures the overall performance approach. This makes it very difficult to learn about the role of different parameters from these results and to build better training pipelines. Furthermore, OCR research typically is conducted in very specific contexts - concerning specific corpora, typefaces, transcription rules and codec sizes - which can make it very tricky to compare single numeric measures like CER across studies.

In this study, we use systematic high-dimensional grid searches on well-defined training and test sets with consistent rules to produce reliable comparisons of different training parameters\footnote{For example, we look at many different combinations of line input and network architecture (see Section \ref{section:research-questions-order}). We do not serialize this search by first looking for an optimal input representation (that works best for one specific network) and then, later, investigating an optimal network for that kind of input.}. Furthermore, we offer detailed ablation studies to understand the role and contribution of each single component, such as individual data augmentation operations. Finally, we offer even more reliable data by sampling model distributions (i.e. retraining models multiple times) and reporting mean and variance for many of our results. The latter is not a common practice yet \cite{berg_TrainingVariancePerformance_2017}.

A very recent study by Clausner et al. on training OCR systems shares various goals with our study \cite{clausner_EfficientEffectiveOCR_2020}. However, Clausner et al.'s focus is on streamlining established tools and configurations and enabling fast training. The authors do not investigate DNN methods, but discuss data augmentation. %

\section{Method}
\label{section:part1-method}

We now set out to investigate the four research questions we stated in Section \ref{section:Deep-OCR}. In the first part of this paper we will tackle the questions one, two and three. The last question -- the influence of the amount of training data on the overall model accuracy -- will be considered in the final part of this study (see Section \ref{section:ablation}). In this final part, we will also present the results of a comprehensive ablation study, which combines all of the previously gathered insights.

For working on the first three questions, we will employ a small annotated corpus sampled from the Berliner Börsen-Zeitung (BBZ) that consists of about 3,000 lines of ground truth. The BBZ is a historical German newspaper published between 1855 and 1944 and is the subject of a current research project we work on together with colleagues from economic history. %
More details about the corpus can be found in Section \ref{section:bbz}.

\subsection{Finding a Good Input Data Representation}
\label{section:line-input-questions}

Line images need to be fed into the network at a specific height that matches the input layer dimensions of the CRNN. This operation involves a resampling of the line image after it was cropped from the scanned image. An important question is which height should be chosen for this process.
A second question relates to the pixel depth of the image. Should the CRNN be given the full grayscale image, or should one convert the line into a monochrome image - as is common in many OCR pipelines \cite{lamiroy_DocumentAnalysisAlgorithm_2011, reul_OCR4allOpenSourceTool_2019} - and then feed the latter to the CRNN? In other words, should the separation of text and background happen inside the CRNN or as a preprocessing step?

Finally, line height and binarization might influence one another: with binarization, a different line height might be optimal than without binarization. Binarization might present a form of aggregation of features that might be better suited for smaller line heights (also see Section \ref{section:research-questions-order}). %

While line height and binarization are two very important aspects we investigate, there are other aspects we neglect in this study. For example, we do not evaluate varying white-space padding, which is a common practice when feeding line images into CRNNs, and which can have an effect on accuracy \cite{martinek_TrainingStrategiesOCR_2019}.
We neither look into the effects of skipping the deskewing step. We performed some preliminary experiments with not-deskewing line images and found that the results were in accordance with what has been reported before, namely that "deep convolutional layers with 1D LSTMs still perform significantly better with [geometric line] normalization than without" \cite{breuel_HighPerformanceOCRPrinted_2013}. Accordingly, we deskew all line input using baseline information detected with Tesseract \cite{smith_OverviewTesseractOCR_2007}.

\subsubsection{Line Height}

Choosing a line height for a CRNN architecture is a tradeoff between getting more detail and getting better generalization.  If a line height is too small, too much information might get lost to differentiate certain characters \cite{tong_MACRNNMultiscaleAttention_2020}. This is especially true for historical typefaces like blackletter where very small details can produce completely different characters. For example \longs, i.e. long s, is a different character than f. Such details, which can be contained in just a few pixels in a high-resolution scan, can get easily lost during resampling to a line height that is too small, thereby breaking the recognition process. The larger the height is on the other hand, the higher the risk gets that the CRNN is not learning generalized features and is overfitting.

We picked the following two line heights based on established best practices found in current OCR and HTR systems:

\paragraph{48 pixels} Of the three state of the art open-source OCR solutions that are used in research and non-research environments\footnote{See \url{https://ocr-d.de/en/models.html}}  Tesseract, Ocropy and Calamari, at least the latter two use a line height of 48 px for their pretrained models \cite{reul_StateArtOptical_2019, wick_CalamariHighPerformanceTensorflowbased_2018, breuel_HighPerformanceText_2017, breuel_HighPerformanceOCRPrinted_2013}. The default line height of Tesseract's CRNN module is not officially documented, however the source code seems to indicate that it is 48 px as well\footnote{See \url{https://github.com/tesseract-ocr/tesseract/blob/d33edbc4b19b794a1f979551f89f083d398abe19/src/lstm/input.cpp#L28}}. It seems safe to say that 48 px is a very common and established line height for current CRNN OCR systems, research architectures \cite{kiss_BrnoMobileOCR_2019}, and performance measurements in the OCR community. Especially Calamari as current state of the art system \cite{wick_ComparisonOCRAccuracy_2018, reul_StateArtOptical_2019} is a reliable baseline for this choice.
\paragraph{64 pixels}  A line height of about 64 px is common in HTR systems. For example, Transkribus HTR+ uses a line height of 64 px \cite{michael_D7HTREngine_2018}, and Wigington et al. report using a line height of 60 px for a German dataset \cite{wigington_DataAugmentationRecognition_2017}. 64 px is also the line height found naturally in many high resolution scans of printed documents, which often makes it the natural line height to use when extracting highest quality line images without any downscaling\footnote{For example, the high resolution BBZ scans we work with have a line height of about 64 px for most body text.}.
\bigskip

We did not investigate other, less common line heights at this point, since the involved training runtimes were already at the limit of our time budget, and we aimed to perform a general evaluation of established large and small line heights. Also, line heights smaller than the established 48 px seem like an attempt to reduce information at a very early stage in the pipeline, and therefore seem questionable to us for the same reasons as binarization, even though the data from Namsyl and Konya might warrant further research for certain typefaces \cite{namysl_EfficientLexiconFreeOCR_2019}. On the other hand, 64 px is already a rather large line height for a non-HTR context which we primarily focus on in this study. Finally, we are not aware of systems that use a line height between 48 px and 64 px.

It is noteworthy in this context that in a different, specifically BBZ-related study, we found that a line height of 56 px yielded slight improvements over 48 px. The reason seems to have been related to single symbols (numerical fractions) that became unreadable at a height of 48 px. This result also verifies our impression that 48 px is a good baseline for "small" line heights when dealing with a variety of complex typefaces.

\subsubsection{Binarization}
\label{section:binarization}

For quite some time, binarization has been an important element of many established OCR pipelines \cite{lamiroy_DocumentAnalysisAlgorithm_2011, reul_OCR4allOpenSourceTool_2019}.
However, the actual usefulness of binarization is not entirely clear. For example, while Mart\'{i}nek et. al underline the positive effect of binarized images over grayscale ones for a German 19th century newspaper \cite{martinek_TrainingStrategiesOCR_2019}, Holley reports that she could not observe a "significant improvement in OCR accuracy between using greyscale or bi-tonal files" \cite{holley_HowGoodCan_2009}.

We want to evaluate whether binarization is useful for line-based CRNN architectures or only reasonable for older engines \cite{smith_OverviewTesseractOCR_2007}. Indeed, the information loss induced in binarization seems questionable with newer line-based CRNN engines, as Yousefi et al. argue\footnote{Unfortunately, their data is supported only by one specific experiment.} \cite{yousefi_BinarizationfreeOCRHistorical_2015}. Similary, Michael et al. report that Transkribus HTR+ performs "a contrast normalization without any binarization of the image" \cite{michael_D7HTREngine_2018}.
To verify whether binarization is still useful with CRNNs, we need to turn to an area of study where binarization should make a clear difference. Since binarization tries to achieve the separation of text from background, the task is trivial for scans of modern material that give a clean separation of dark ink and white background. It is therefore of little consequence to investigate the use of binarization by focusing only on modern documents. We should instead turn to older material. Rani et al. note: "In old document images in the presence of degradations (ink bleed, stains, smear, non-uniform illumination, low contrast, etc.) the separation of foreground and background becomes a challenging task. Most of the existing binarization techniques can handle only a subset of these degradations" \cite{rani_NewBinarizationMethod_2019}. To answer how effective binarization is at performing its intended use - namely separating text from background - it is crucial to look at difficult examples instead of simple ones. We specifically focus on historical newspapers, since they are an ideal testing ground that exposes all problems that binarization aims to solve.

There are numerous algorithms for binarization  which can be classified broadly into local (e.g. Sauvola) and global (e.g. Otsu) approaches. However, opinions differ as to which approaches achieve the best results. To pick a baseline against which we can test, i.e. to compare binarized and non-binarized approaches, we need to decide on at least one specific, reasonably good and established binarization approach. 
A short review of possible candidates includes the following: Gupta et al. report that "the underlying assumption behind the Otsu method appears to best model the truth behind historical printed documents, based on results over a broad range of historical newspapers" \cite{gupta_OCRBinarizationImage_2007}. On the other hand, Antonacopoulos and Castilla note that "in historical documents, it is somewhat obvious that global methods are not appropriate due to the non-uniformity of the text and the background" \cite{antonacopoulos_FlexibleTextRecovery_2006}. Ntirogiannis et al. study 8 binarization algorithms, with Otsu's performance never being among the top-4 ranked algorithms, whereas Sauvola for example is always in the top-3 \cite{ntirogiannis_PerformanceEvaluationMethodology_2013}. Therefore, and in accordance with our own experiments, we pick Sauvola as baseline binarization algorithm for the evaluations in the coming sections, since it generally gives a very reasonable performance.

We are aware that even more algorithms exist, for example, Lund et al. find that multi-threshold binarization yields best results for 19th century newspapers \cite{lund_CombiningMultipleThresholding_2013}. A summary of other recent binarization improvements is found in \cite{ahn_TextlineDetectionDegraded_2017}. Very recently Barren showed how to combine various classical binarization algorithms into a new framework that seems to outperform some DNN approaches\footnote{\url{https://arxiv.org/abs/2007.07350}}.

\subsubsection{Investigated Combinations}

To better understand the interplay between line height and binarization, we investigate four configurations of the two parameters (see Table \ref{tab:limg}).

\begin{table}[htb]
\center
\begin{tabular}{l c c} 
Name & Line height in pixels & Binarization\\
\hline
\emph{gray48} & 48 & false\\
\emph{bin48} & 48 & true\\
\emph{gray64} & 64 & false\\
\emph{bin64} & 64 & true\\
\end{tabular}
\caption{\label{tab:limg}Line image configurations.}
\end{table}
Binarization is performed by applying a Sauvola filter \cite{sauvola_AdaptiveDocumentImage_2000} with a window size of $\frac{h}{2} - 1$.
Deskewing and downscaling is implemented using OpenCV's \textproc{INTER\_AREA} filter\footnote{\url{https://opencv.org/}} for both 48 px and 64 px  versions.

In order to not lose additional information, we only scale line images vertically, but not horizontally, i.e. we do not scale according to aspect ratio.

\subsection{Finding a Good Network Architecture}
\label{section:crnn-arch}

As discussed in the related work overview, different OCR engines employ a variety of CRNN architectures, but it is not an easy question to answer which influence these architectures have on the overall OCR performance. A system like Calamari allows the user to specify a large range of CRNN architectures, but in the absence of best practices on which architectures work better than others, most users stick with the defaults.

To get a better understanding of the role and performance of different architectures, this section explores a method to evaluate a larger subset of network architectures. To find an optimal architecture we use a grid search on neural architectures, which we call Neural Architecture Grid Search (NAGS). We avoid the established term Neural Architecture Search (NAS) for this task. The latter term refers to much more sophisticated methods than grid searches \cite{zoph_NeuralArchitectureSearch_2017, elsken_NeuralArchitectureSearch_2019} such as, for example, Population Based Training \cite{liaw_TuneResearchPlatform_2018, li_GeneralizedFrameworkPopulation_2019}.

We first define a framework for generating a search space of good network architectures suitable for OCR in Section \ref{section:framework-crnn}. We then describe an approach of searching inside that space in Section \ref{section:finding-psi} to find an optimal architecture we call $\Psi$.

\subsubsection{A Framework for CRNN Architectures}
\label{section:framework-crnn}

\begin{table}[htb]
\center
\begin{tabular}{l l l l l l l}
& layer & kernel size & stride & \#filters & \#filters & \#units \\
& & & & \cite{wick_CalamariHighPerformanceTensorflowbased_2018} & \cite{wick_CalamariOCR_2018} & \\
\hline
1 & $conv_1$ & $3 \times 3$ & $1 \times 1$ & 64 & 40\\
2 & $pool_1$ & $2 \times 2$ & $2 \times 2$ & \\
3 & $conv_2$ & $3 \times 3$ & $1 \times 1$ & 128 & 60\\
4 & $pool_2$ & $2 \times 2$ & $2 \times 2$ & \\
5 & $lstm$ & & & & & 200
\end{tabular}
\caption{\label{tab:calamari-arch}Calamari's default network architectures from \cite{wick_CalamariHighPerformanceTensorflowbased_2018} and \cite{wick_CalamariOCR_2018} (without dropout, dense and softmax layers).}
\end{table}
Calamari's default architecture\footnote{The defaults in the current implementation \cite{wick_CalamariOCR_2018} differ from what is used in the paper\cite{wick_CalamariHighPerformanceTensorflowbased_2018}. \cite{wick_CalamariHighPerformanceTensorflowbased_2018} describes two CNN layers with 64 and 128 filters, whereas, as of April 2020, the implementation \cite{wick_CalamariOCR_2018}, uses 60 and 120 filters by default. We show both sets of values since both seem interesting.} is shown in Table \ref{tab:calamari-arch}. Note that we ignore dropout, dense and softmax layers at the end of Calamari's network and assume them as granted for now.

We try to generalize from Calamari's default network architecture and define a framework for generating other interesting architectures. To help us decide which of the various parameters of the Calamari network we should vary, we look at recent CRNN architectures for text recognition like \cite{breuel_HighPerformanceOCRPrinted_2013, wick_CalamariHighPerformanceTensorflowbased_2018, michael_D7HTREngine_2018, martinek_BuildingEfficientOCR_2020, namysl_EfficientLexiconFreeOCR_2019, mehrotra_CollaborativeDeepNeural_2019, zhou_EASTEfficientAccurate_2017}, established CNN architectures such as VGG \cite{simonyan_VeryDeepConvolutional_2015}, and turn to three pieces of advice from the "General Design Principles“" in \cite{szegedy_RethinkingInceptionArchitecture_2016}:

\begin{itemize}
    \item [(\textbf{G1})] "Balance the width and depth of the network. Optimal performance of the network can be reached by balancing the number of filters per stage and the depth of the network. Increasing both the width and the depth of the network can contribute to higher quality networks." \cite{szegedy_RethinkingInceptionArchitecture_2016}
    \item [(\textbf{G2)}] "Avoid representational bottlenecks, especially early in the network." \cite{szegedy_RethinkingInceptionArchitecture_2016} 
    \item [(\textbf{G3)}] "[C]onvolutions with filters larger $3 \times 3$ a might not be generally useful as they can always be reduced into a sequence of $3 \times 3$ convolutional layers" \cite{szegedy_RethinkingInceptionArchitecture_2016}
\end{itemize}
For our study, we derive the following consequences from those design principles:

\begin{itemize}
    \item From (G1) -- We vary both the number of convolution layers (depth) as well as the number of filters (width).
    \item From (G2) -- We try to keep pooling layers late in the network or omit them altogether. Another view of this guideline is that we replace pooling layers with convolutional layers, as proposed by Springenberg et al. \cite{springenberg_StrivingSimplicityAll_2015}. For simplicity, we do not investigate replacing pooling layers with larger strides though, as also proposed by Springenberg et al. \cite{springenberg_StrivingSimplicityAll_2015}. In fact, we keep strides as small as possible ($1 \times 1$ for convolution layers, $2 \times 2$ for pooling layers).
    \item From (G3) -- We ignore convolution kernel sizes larger than $3 \times 3$ and focus on depth to simulate larger kernel sizes (this also follows results from \cite{simonyan_VeryDeepConvolutional_2015}). As for smaller kernel sizes, we ignore $1 \times 1$ and asymmetric convolutions for the sake of simplicity. The effects of the latter (see \cite{szegedy_RethinkingInceptionArchitecture_2016, springenberg_StrivingSimplicityAll_2015}) are beyond the scope of this study.
\end{itemize}
Based on these assumptions, we define a specific network architecture with four parameters $N$, $R$, $K$ and $P$:

\begin{itemize}
    \item $K$ and $P$ model the \emph{depth} of the network by indicating that the network should have $K$ convolution layers and $P$ pooling layers.
    \item $N$ and $R$ model the \emph{width} of the network by defining how many filters the convolution layers will get. Higher values of $N$ and $R$ will increase the overall width.
\end{itemize}
Specifically, the i-th convolution layer will obtain $f_i$ filters, where $f_i$ is defined as

\[
f_K = N \quad \cap \quad  f_i = max(\lfloor \frac{f_{i+1}}{R} \rfloor, 8)
\]
In other words, $N$ is the (maximum) number of filters $N$ in the latest convolution layer. For each convolution layer going towards the input, this number will decrease by $R$.
We add one more parameter $M$ to model the number of LSTM cells and their units that each CRNN architecture for text recognition employs at the end of the network in one way or another \cite{breuel_HighPerformanceOCRPrinted_2013}. Using $M$, we are also able to model stacked LSTMs, which is an older idea \cite{graves_SpeechRecognitionDeep_2013}, but is used in recent OCR models \cite{michael_D7HTREngine_2018}. We do not investigate the use of newer propositions like BLSTMs, MDLSTMs, and nested LSTMS \cite{bukhari_InvestigativeAnalysisDifferent_2018, moniz_NestedLSTMs_2017}.

\begin{table}[htb]
\center
\begin{tabular}{l | l l l l}
    portion & layer & kernel size & stride & \#filters/units \\
	\hline
	\\
	CNN & $conv_1$ & $3 \times 3$ & $1 \times 1$ & $f_1$ \\
	& $conv_2$ & $3 \times 3$ & $1 \times 1$ & $f_2$ \\
	&... \\
	& $conv_{K-1}$ & $3 \times 3$ & $1 \times 1$ & $f_{K-1}$ \\
	& $pool_{P-1}$ & $2 \times 2$ & $2 \times 2$ & \\
	& $conv_{K}$ & $3 \times 3$ & $1 \times 1$ & $f_K$ \\
	& $pool_{P}$ & $2 \times 2$ & $2 \times 2$ & \\
	\\
	LSTM & $lstm$ & & & $M_1$\\
	&...\\
	& $lstm$ & & & $M_{|M|}$\\
\end{tabular}
\caption{\label{tab:arch}Structure of investigated architectures.}
\end{table}

The network layout derived from one set of $N$, $R$, $K$, $P$ and $M$ is shown in Table \ref{tab:arch}. 
Note that we split the CRNN network into a an CNN portion (everything up to the LSTM modules) and a LSTM portion (consisting only of the LSTM modules at the very end).
Calamari's default architecture from \cite{wick_CalamariHighPerformanceTensorflowbased_2018} would be described as $(N=128, R=2, K=2, P=2, M=\{200\})$. 

\subsubsection{Findings $\Psi$}
\label{section:finding-psi}

We now look for an optimal CRNN architecture which we call $\Psi$. For $\Psi$, we keep $M$ at the Calamari default of $\{200\}$, and focus solely on changing the CNN portion of the network by varying $N$, $R$, $K$ and $P$. We will turn to the LSTM portion and investigate the effects of changing $M$ in later sections.

\begin{table}[htb]
\center
\begin{tabular}{l l l}

& $R=1.5$ & $R=2$ \\
\hline
narrow & 64 & 64 \\
medium & 124 & 128 \\
wide & 240 & 256 \\

\end{tabular}
\caption{\label{tab:widths}Values of $N$ for different $R$s.}
\end{table}
We investigate $R=1.5$ (default from \cite{wick_CalamariOCR_2018}) and $R=2$ (i.e. power of 2, which is a very common CRNN architecture default \cite{wick_CalamariHighPerformanceTensorflowbased_2018, breuel_HighPerformanceOCRPrinted_2013, michael_D7HTREngine_2018}, as well as in other common CNN architectures, e.g. ResNet \cite{he_DeepResidualLearning_2016}).
For $N$ we investigate the three variants shown in Table \ref{tab:widths}. For $R=1.5$ and by convention, we chose $N$ such that we obtain even filters counts $f_K$ for all $K$.

Following (G1), we investigated all values between 2 (Calamari default) and 15 for $K$. 
Following (G2), we used $P=2$ (Calamari default) and $P=1$, but not $P=3$. We omitted $P=0$ because preliminary results showed that this did not work well at all.
Following (G3) and the fact that at least 10 layers have been used in fully convolutional networks for text recognition \cite{namysl_EfficientLexiconFreeOCR_2019}, we focus more on depth ($K$) than on width ($N$ and $R$). This is now a general practice in deep learning, as the "depth of a neural network is exponentially more valuable than the width of a neural network" \cite{wang_OriginDeepLearning_2017}. 
To investigate $M$, we first assume $M={u}$, i.e. a single LSTM cell, and look at the influence of changing $u$, the number of units in this cell. Number between $u=200$ and $u=256$ are very common \cite{wick_CalamariHighPerformanceTensorflowbased_2018, namysl_EfficientLexiconFreeOCR_2019, michael_D7HTREngine_2018}. We will investigate $u$ for $150 \leq u \leq 500$.
We will also investigate the effect of stacking several LSTM modules as done in \cite{michael_D7HTREngine_2018}.

\subsection{Finding the Right Data Augmentation}
\label{section:data-augmentation-questions}

Data augmentation is an important procedure in deep learning to avoid overfitting and help generalization; it was already applied in LeNet-5 as "one of the first applications of CNNs on handwritten digit classification" \cite{shorten_SurveyImageData_2019}. Transferring this statement to the area of OCR, the question arises: which data augmentation should be considered for current CRNN architectures and how should it be configured exactly?
In practice, there are many aspects of data augmentation related to general deep learning and CNNs \cite{shorten_SurveyImageData_2019, taylor_ImprovingDeepLearning_2018, devries_ImprovedRegularizationConvolutional_2017} or OCR respectively \cite{simard_BestPracticesConvolutional_2003}. Surveying the field, we find that many augmentations, such as color space transforms, flipping and cropping, do not make sense for OCR data. As detailed below, we investigate two fields of data augmentation that %
are commonly used in OCR: (A) geometric deformations and (B) image degradation.

We do not investigate newer approaches of automatically finding  classes of optimal augmentations \cite{cubuk_AutoAugmentLearningAugmentation_2019}.
The augmentation studies in this section will use a 100\% ratio, i.e. for each line image in our training set we generate one additional augmented image.

\subsubsection{Geometric Deformations}
\label{section:geometric-deformations}

In the context of OCR, geometric deformations might help a neural network to learn effects like warped paper and uneven printing. 
Geometric deformation algorithms can be classified as linear (or affine) and non-linear transformations. The latter generally seem to work better. For example, \cite{simard_BestPracticesConvolutional_2003} note that "[a]ffine distortions greatly improved our results," but "our best results were obtained when we used elastic deformations." Similarly, \cite{namysl_EfficientLexiconFreeOCR_2019} report "that non-linear distortions can [...] reduce the error rate of models".
Accordingly, we take a closer look at two forms of non-linear distortions that we apply on the whole line image (not single characters):

\begin{itemize}
\item Using a high-resolution grid deformed by Gaussian noise, which is basically the elastic transform described by \cite{simard_BestPracticesConvolutional_2003} and has also been reported by Wigington et al. \cite{wigington_DataAugmentationRecognition_2017} in the context of HTR. One implementation is found in ocrodeg\footnote{\url{https://github.com/NVlabs/ocrodeg}}. We will look at various settings of ocrodeg and how they influence the model's CER. To be specific, we will use ocrodeg's \textproc{bounded\_gaussian\_noise} and \textproc{distort\_with\_noise} functions that are controlled by two parameters: $sigma$ is the standard deviation of a Gaussian kernel which is applied to random noise, $maxdelta$ is a scaling factor modelling the strength of the overall distortion\footnote{We used the ocrodeg version currently shipping with Calamari, i.e. git hash 728ec11. It shows small but probably negligable differences to the official ocrodeg repository like switching from pylab to numpy.}.

\item Using a low-resolution grid deformed by uniform noise. Again, we will investigate the effect of various settings, such as grid size and amount of distortion.
\end{itemize}

\subsubsection{Image Degradation and Cutout}
\label{section:image-degradation}

Image degradation can help a network to learn the effects of missing ink, spreading of ink, smudges, ink bleed-through, and other forms of paper degradation that are all very common in historical documents. By leaving out information, it can also further regularization \cite{devries_ImprovedRegularizationConvolutional_2017}. We investigate three types:

\begin{itemize}
    \item Adding Gaussian noise, which "can help CNNs learn more robust features" \cite{shorten_SurveyImageData_2019}. We investigate the effect of applying various noise distributions.

    \item Changing the contrast and brightness of the line images. Specifically, we apply some random contrast and brightness, but limit the maximum for each by two parameters. We rely on the \textproc{RandomBrightnessContrast} implementation of the \emph{albumentations} library \cite{buslaev_AlbumentationsFastFlexible_2020} and turn off \textproc{brightness\_by\_max}.
    
    \item Adding random blobs that imitate ink or paper and in fact delete pieces of information. Again, we investigate different distributions. This kind of augmentation is sometimes called "cutout" or "random erasing" and is known to have "produced one of the highest accuracies on the CIFAR-10 dataset" \cite{shorten_SurveyImageData_2019, devries_ImprovedRegularizationConvolutional_2017}.
    We use ocrodeg's \textproc{random\_blobs} method to add such blobs, which offers a versatile approach to generate various blob configurations, which are guided by two parameters called \emph{amount} and \emph{size}. The parameter \emph{amount} (called \emph{fgblobs} in ocrodeg) models the amount of blobs as fraction of the total number of pixels in a line image. The parameter \emph{scale} (\emph{fgscale}) models the size of the blobs.
    Ocrodeg assumes bitonal input for its blotch augmentation and it adds blotches that are pure white or black. This is geared towards binarized input, but a highly unrealistic augmentation for grayscale input, which we also want to evaluate. We therefore modify ocrodeg to use a $0.05$ brightness quantile instead of pure black for background blotches and a $0.75$ brightness quantile instead of pure white for foreground blotches.
\end{itemize}

\subsection{Technical Details of the Evaluation Procedure}

\subsubsection{Order of Evaluation}
\label{section:research-questions-order}

A practical problem with evaluating the influence of various factors such as line height,  binarization, the impact of network architectures and finally data augmentation are their interdependencies and complex interactions.
For example, it is quite possible that we find one network architecture that works great with non-binarized tall images, while another one works great with binarized images of low height. %
Therefore we evaluate the impact of line input and binarization not as a separate step, but as part of the NAGS, by introducing the line height and binarization parameters into the NAGS and dealing with them similar to $R$ and $N$ (see Table \ref{tab:widths}).
For data augmentation on the other hand, we assume that every good network architecture will benefit in similar ways from good data augmentation. Therefore we evaluate various options of data augmentations only on our optimal architecture $\Psi$. The latter is also a necessary decision to deal with combinatorial explosion and keep the NAGS search times within computational constraints.

\subsubsection{Measuring Model Performance}
\label{section:model-dist}

\begin{figure}[htb!]
    \centering
    \includegraphics[scale=0.9]{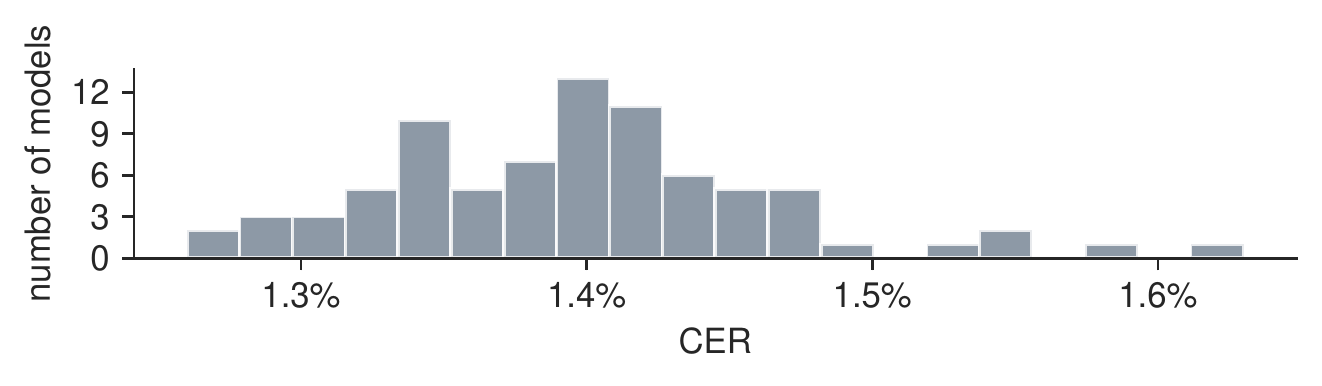}
    \caption{Distribution of CERs of 81 identical training setups using binary identical input data (binarized, deskewed lines with height 48) and Calamari default parameters. Trained on TensorFlow 2.0.0 using fixed random seeds for Calamari, Python, numpy and TensorFlow. $\mu CER=1.40\%$, $\sigma CER=0.07$, $min$ $CER = 1.26\%$, $max$ $CER = 1.63\%$.}
    \label{fig:calamari-samples}
\end{figure}

It is well known that many implementations of DNNs on GPUs have non-deterministic components during training\footnote{For example, see \url{https://github.com/NVIDIA/tensorflow-determinism}}. This is illustrated in Figure \ref{fig:calamari-samples}.

The mean CER of the 81 models shown in Figure \ref{fig:calamari-samples} is $1.4\%$, but the CERs of single models vary between $1.26\%$ and $1.63\%$. This volatility of results makes objective comparisons of models unsound if CER differences are small and only few models are trained. For example, assuming a normal distribution and $\sigma$ from Figure \ref{fig:calamari-samples}, 2 model trainings are needed differentiate a CER percentage difference of $0.1$ with $95\%$ confidence. However, 8 trainings are needed to detect a CER percentage difference of $0.05\%$ with the same confidence.

While this issue is well known among practitioners, there seems to exist little research on how to deal with it. One of the few studies to clearly state the problem \cite{berg_TrainingVariancePerformance_2017} notes: "it is still common practice to report results without any information on the variance".

To compare the performance of two models, \cite{berg_TrainingVariancePerformance_2017} describe three approaches: (1) "[c]ompare the best possible results" \cite{berg_TrainingVariancePerformance_2017}, which "really depends on the probability mass contained in the lower tail; if this is small it will be very difficult to sample" \cite{berg_TrainingVariancePerformance_2017}. (2) compare the means, (3) "[d]etermine the probability that a randomly sampled models of one method is better than that the other." (3) is arguably the best approach, but beyond time and technical limits in some cases. In this study, we generally avoid (1), but - due to computational constraints - are not following (3) rigorously either. Still, in many cases our sample sizes should allow reasonable confidence. Also, our grid searches that compute many models with only very small differences should allow us to detect structures and clusters of high-performing and similar models, even with low sample sizes like 2 or 3.

\subsubsection{Technical Platform}

As a technical platform for our study we use Calamari \cite{wick_CalamariHighPerformanceTensorflowbased_2018, wick_CalamariOCR_2018}. As opposed to other software (e.g. Tesseract or Transcribus), Calamari has some major advantages for our study:%

\begin{itemize}
\item GPU-support through TensorFlow, which is a big advantage for the huge training computation load in this study
\item Calamari offers a well-thought-out interface for specifying custom network architectures and is open source
\item Calamari implements voting to further improve quality of predictions
\end{itemize}

\subsubsection{Training Corpus}
\label{section:bbz}

For the evaluation of the first three research questions, we employ a small but diverse corpus of 2,807 lines, which consists of 90\% blackletter and 10\% Antiqua\footnote{The ground truth was manually transcribed from the BBZ. The lines were sampled from the years 1872, 1885, 1897, 1902, 1918, 1925 and 1930.}. Although the corpus is rather small, it exposes many features that are challenging for OCR models and are typical of historical newspapers: degraded paper quality, mixing of different typefaces (Antiqua and German Fraktur), very small print, and various special symbols (e.g. for various currencies). The German Fraktur typeface poses a specific set of difficulties for OCR
\cite{vorbach_AnalysenUndHeuristiken_2014}. The corpus furthermore contains special header typefaces, a variety of typefaces for numbers, numerical fractions (e.g. 1/2, 7/8 and so on), bold text and finally text with wide letter-spacing (\emph{Sperrsatz}). We use 20\% of the data as validation data and the remaining 80\% as training data. Due to the small size of the corpus, we do not use a dedicated test set, but rely on the validation set as measurement of accuracy for our experiments in part 1 of this study. We will use a large independent test set in part 2 of this study.

\section{Results}
\label{section:part1-results}

This section presents results for the previously introduced OCR components  when used on our training corpus.

\paragraph{Some preliminary Remarks on the CER:} For computing CERs, we use the "mean normalized label error rate" from Calamari's \textproc{calamari-eval} script. To be exact, if there are $n$ lines, and $L_i$ is the Levenshtein distance between ground truth and prediction for line $i$, and $C_i$ is the number of characters in line $i$, then Calamari computes $\frac{\sum_{i=1}^{n} L_i}{\sum_{i=1}^{n} C_i}$, but not $\frac{\sum_{i=1}^{n} \frac{L_i}{C_i}}{n}$, i.e. a line's contribution to the total CER depends on its length.

\subsection{Findings on $\Psi$ and $I$}

\subsubsection{Non-binarized input with less pooling layers works best}
\label{section:non-binarized-works-best}

\begin{figure}[ht!]
    \centering
    \includegraphics[scale=0.85]{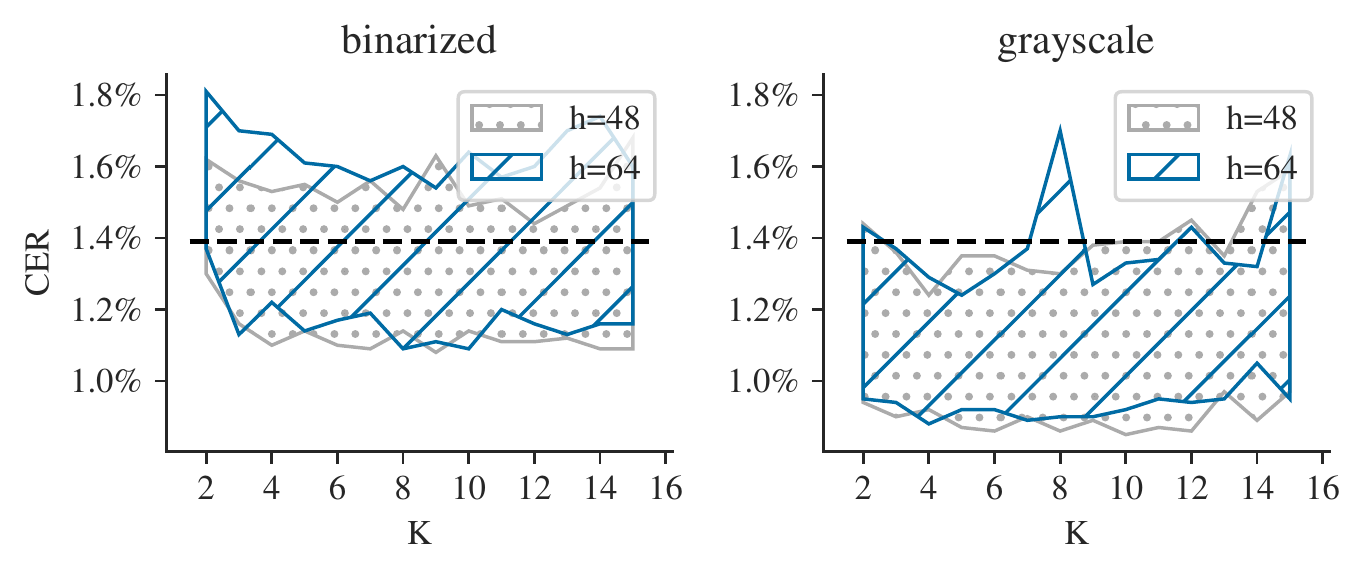}
    \caption{Minimum and maximum CER of trained models by line height used. Lower CERs are better.}
    \label{fig:bigpic}
\end{figure}

We first present two central findings on binarization and the number of pooling layers in the network architecture, which we termed $P$ earlier (see Section \ref{section:crnn-arch}).
Figure \ref{fig:bigpic} shows the CERs of a wide range of network models trained on our investigated two line heights. For each $K$ (on the x-axis), we see the best and worst CERs obtained (best and worst in all investigated $N$, $R$, and $P$ as described in Section \ref{section:finding-psi}). 
We found that the best grayscale models (right) outperform the best binarized models (left) for both investigated line heights and across all investigated network depths ($K$). We do not see a big difference in the performance between line heights, though it seems as if line height 64 never outperforms line height 48.

\begin{figure}[htb!]
    \centering
    \includegraphics[scale=0.85]{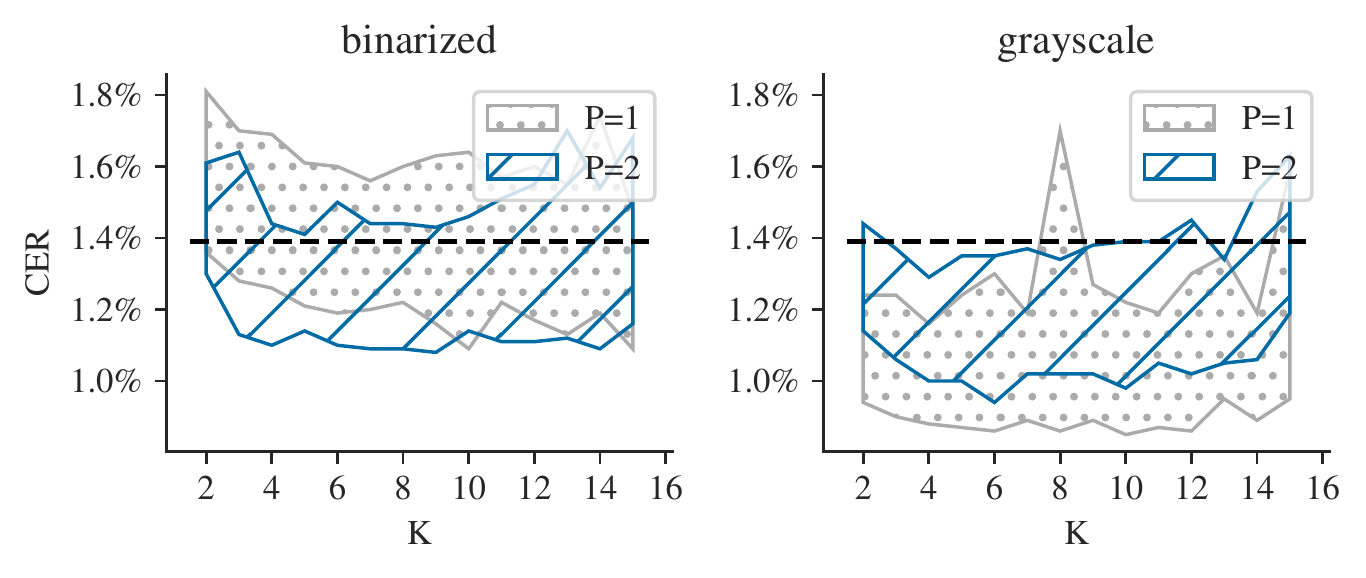}
    \caption{Minimum and maximum CER of trained models by number of pooling layers $P$. Lower CERs are better.}
    \label{fig:pooling}
\end{figure}

Figure \ref{fig:pooling} shows the influence of the number of pooling layers $P$. For binarized input (left), the best models utilize 2 pooling layers (blue dashed area), whereas for grayscale input (right), the best models utilize only 1 pooling layer (gray dotted area). This seems to be a confirmation of the hypothesis stated in Section \ref{section:binarization}, that an additional pooling layer %
might lose detailed and useful information in the grayscale case. Notably, the best $P=2$ grayscale models perform better than any binarized images (compare dotted area right to blue area left).

\subsubsection{Results of Neural Architecture Search}
\label{section:results-nas}

We now present an overall view of the large combined input and architecture search we performed. As described at the beginning of Section \ref{section:part1-method}, we present a detailed picture of the dependencies of line input format - line height and binarization - and various aspects of network architecture - namely $N$, $R$ and $K$ from Section \ref{section:crnn-arch}.

\begin{figure}
    \centering
    \includegraphics[scale=0.6]{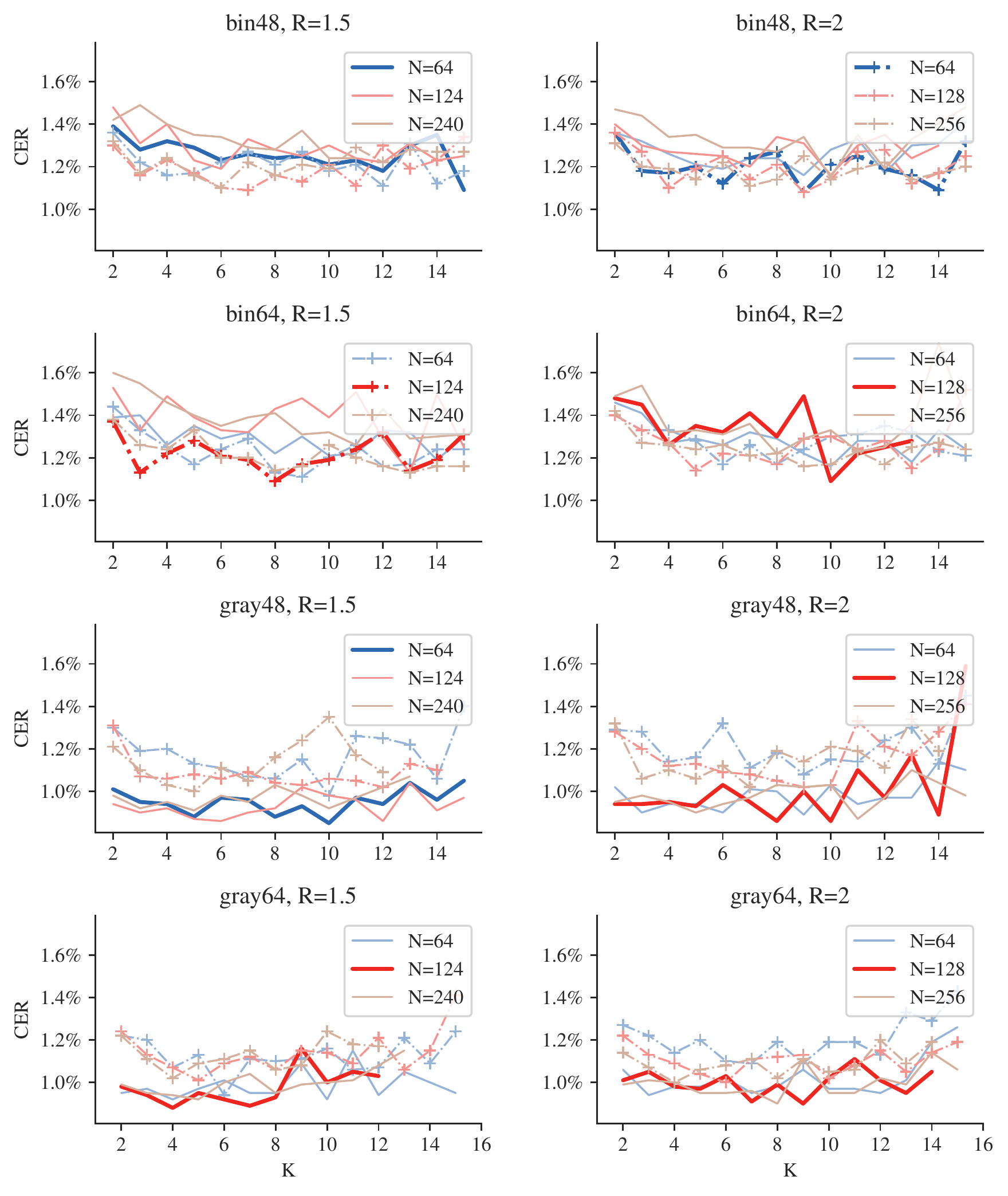}
    \caption{Minimum CERs in percent achieved by various network architectures. Left: $R$ = 1.5. Right: $R$ = 2. Top to bottom: $I$=bin48, $I$=bin64, $I$=gray48, $I$=gray64. $N$: see legends. $P$ is included for all tested values, but not explicitly shown.}
    \label{fig:best-arch}
\end{figure}

Figure \ref{fig:best-arch} shows detailed results of the NAGS for all the parameters that were taken into account. Although the data is noisy due to the low sampling size and large search space, the superior performance of grayscale (lower four plots) vs. binarized modes (upper four plots) is evident.

Table \ref{tab:best_nas_min} shows the top 5 architectures by \emph{min} CER, and Table \ref{tab:best_nas_mean} by mean CER. 
$I=$\emph{gray48} is dominant, as are architectures that have \emph{medium} network width (see Table \ref{tab:widths}). 
For the latter reason, we pick the smallest such architecture for further investigation (i.e. for $\Psi$), which is \#2 from Table \ref{tab:best_nas_min} ($N=124$, $R=1.5$, $K=6$), and avoid the narrow network architecture \#1. Note that ($N=124$, $R=1.5$, $K=12$) looks like an even better pick as it occurs in the top 3 in both Tables \ref{tab:best_nas_min} and \ref{tab:best_nas_mean}. We will investigate that specific model later in section \ref{section:stage8-fine-tuning}.

\begin{figure}
    \centering
    \includegraphics[scale=0.9]{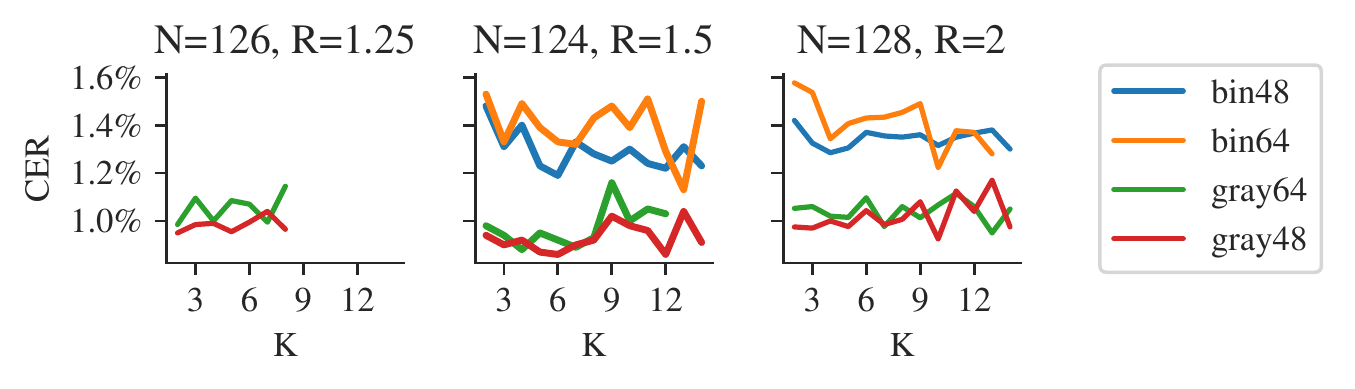}
    \caption{Overview of CERs of network architectures using \emph{medium} network width (see Table \ref{tab:widths}) with varying $R$, $K$ and $I$.}
    \label{fig:best-arch-pick}
\end{figure}

For now, we stick with $K=6$. Figure \ref{fig:best-arch-pick} gives an overview of \emph{medium} network width architectures, which consistently give the most promising results. The two mentioned configurations ($K=6$ and $K=12$) are seen in the valleys of the red line in the middle panel.

\begin{table}[htb!]
\center
\begin{tabular}{c c c c c c c} 
\# & $I$ & $N$ & $R$ & $K$ & $P$ & min CER\\
\hline
1 & gray48 & 64 & 1.5 & 10 & 1 & \textbf{0.85\%} \\
2 & gray48 & 124 & 1.5 & 6 & 1 & 0.86\% \\
3 & gray48 & 124 & 1.5 & 12 & 1 & 0.86\% \\
4 & gray48 & 128 & 2 & 8 & 1 & 0.86\% \\
5 & gray48 & 128 & 2 & 10 & 1 & 0.86\% \\
\end{tabular}
\caption{\label{tab:best_nas_min}Best Architectures from NAGS when sorted by the best CER obtained (i.e. min CER) in multiple runs.}
\end{table}

\begin{table}[htb!]
\center
\begin{tabular}{c c c c c c c} 
\# & $I$ & $N$ & $R$ & $K$ & $P$ & mean CER\\
\hline
1 & gray48 & 256 & 2 & 11 & 1 & \textbf{0.90\%} \\
2 & gray48 & 124 & 1.5 & 12 & 1 & 0.91\% \\
3 & gray48 & 128 & 2 & 10 & 1 & 0.93\% \\
4 & gray48 & 64 & 1.5 & 8 & 1 & 0.94\% \\
5 & gray64 & 64 & 1.5 & 10 & 1 & 0.94\% \\
\end{tabular}
\caption{\label{tab:best_nas_mean}Best Architectures from NAGS when sorted by the mean CER obtained in multiple runs.}
\end{table}

\subsubsection{Further Improvement of $\Psi$}

\begin{figure}[htb!]
    \centering
    \includegraphics[scale=0.85]{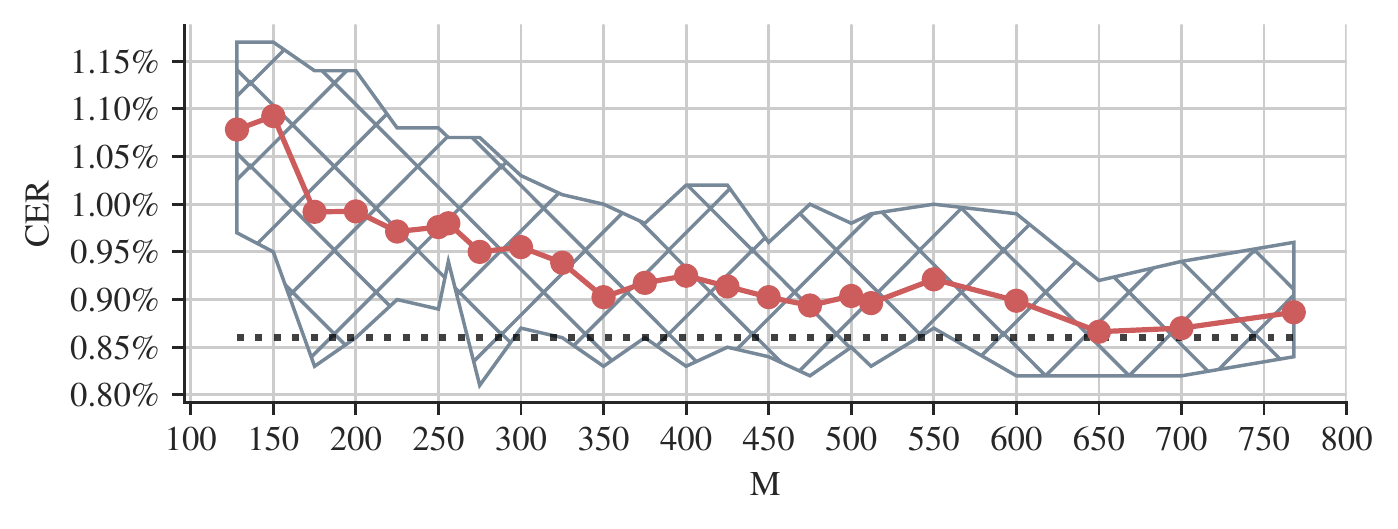}
    \caption{$\Psi$ with one LSTM cell with a varying number of units. Thick line shows mean ($n >= 6$). Dotted line shows min CER with $\Psi$ and $M=\{200\}$.}
    \label{fig:lstm1}
\end{figure}

Based on $\Psi$, we experimented with one LSTM cell and varying number of its units (see Table \ref{fig:lstm1}). We find the best mean CER results at $M=\{650\}$ and $M=\{700\}$, which both show a mean CER improvement of about $0.13\%$ (compared to $M=\{200\}$). Both produce a mean CER that is roughly the min CER seen with $M=\{200\}$.

We also experimented with stacking LSTMs, but saw dramatic worsening of performance to a mean CER of $1.9\%$ for two ($n=15$), and $8.6$ for three ($n=2$) stacked LSTMs - as opposed to a CER of $0.99\%$ for our default $\Psi$. We did not pursue this further, but suspect flaws with the way we set up the network architecture.
Finally, we trained $\Psi$ with various dropout rates and found that a dropout of $0.5$ works best (see Figure \ref{tab:dropout}). Even though dropout $0.6$ and $0.7$ show the same mean CER, $min$ CER is lowest with $0.5\%$. This result is in accordance with previous results from Srivastava et al., who report that a dropout of value $0.5$ "seems to be close to optimal for a wide range of networks and tasks" \cite{srivastava_DropoutSimpleWay_2014}

\begin{table}[htb!]
\center
\begin{tabular}{c c c c c c} 
dropout & $\mu$ CER & $\sigma$ CER & $min$ CER & $n$ \\
\hline
0.2 & 1.11\% & 0.04 & 1.03\% & 8 \\
0.3 & 1.06\% & 0.07 & 0.95\% & 16 \\
0.4 & 1.03\% & 0.06 & 0.91\% & 13 \\
0.5 & \textbf{0.99\%} & 0.07 & \textbf{0.86\%} & 32 \\
0.6 & \textbf{0.99\%} & 0.06 & 0.88\% & 24 \\
0.7 & \textbf{0.99\%} & 0.06 & 0.89\% & 24 \\
0.8 & 1.03\% & 0.03 & 0.97\% & 16 \\
\end{tabular}
\caption{\label{tab:dropout}CER when training $\psi$ with various dropout rates. The column \emph{n} indicates the number of trained models (i.e. sample size).}
\end{table}

\subsection{Data Augmentation}
\label{section:results-of-data-augmentation}

In this section, we report the results of our analysis of data augmentation. Except for low-resolution distortion, which was done on an earlier architecture, all of these studies were performed on $\Psi$ with $M=\{200\}$.

\subsubsection{Geometric Distortions}

\begin{figure}[htb!]
    \centering
    \includegraphics[scale=0.65]{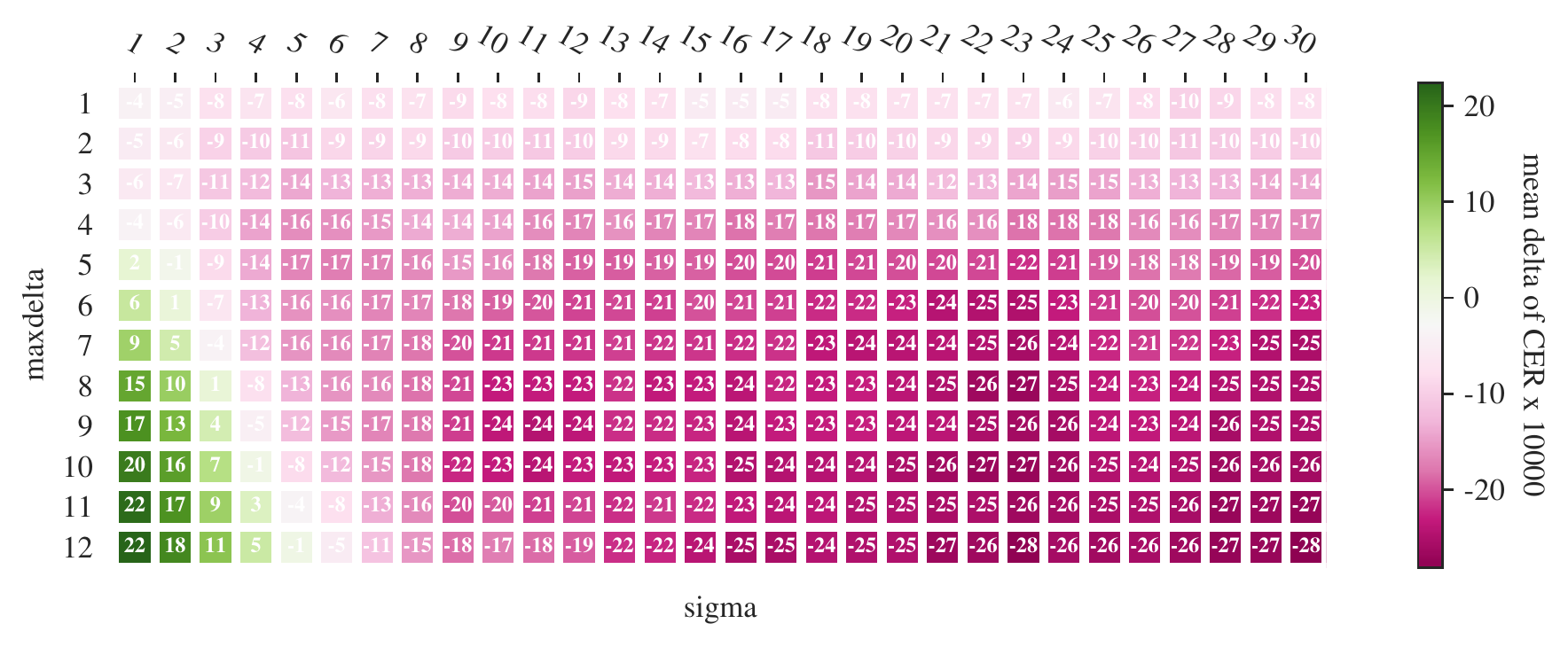}
    \caption{Mean improvement of CER for high-resolution grid distortion using ocrodeg. Neighbouring cells are included in a cell's mean values to increase sample size. A CER delta of $1\%$ is displayed as 100.}
    \label{fig:distort-hires}
\end{figure}

As described in Section \ref{section:geometric-deformations}, we first look at high-resolution grid deformations by investigating the influence of $sigma$ and $maxdelta$ to ocrodeg's function \textproc{bounded\_gaussian\_noise}. Our results are shown in Figure \ref{fig:distort-hires}. Contrary to common practice (the ocrodeg manual's example code sets $sigma$ between 1 and 20 and $maxdelta$ to $5$), it seems that setting these values quite high is beneficial. With $maxdelta=12$ and $sigma=30$ we observe an CER percentage drop of $0.28$. Our study lacks an investigation of even higher parameter values that might yield an even further improvement of CER before dropping off.

\begin{figure}[htb!]
    \centering
    \includegraphics[scale=0.8]{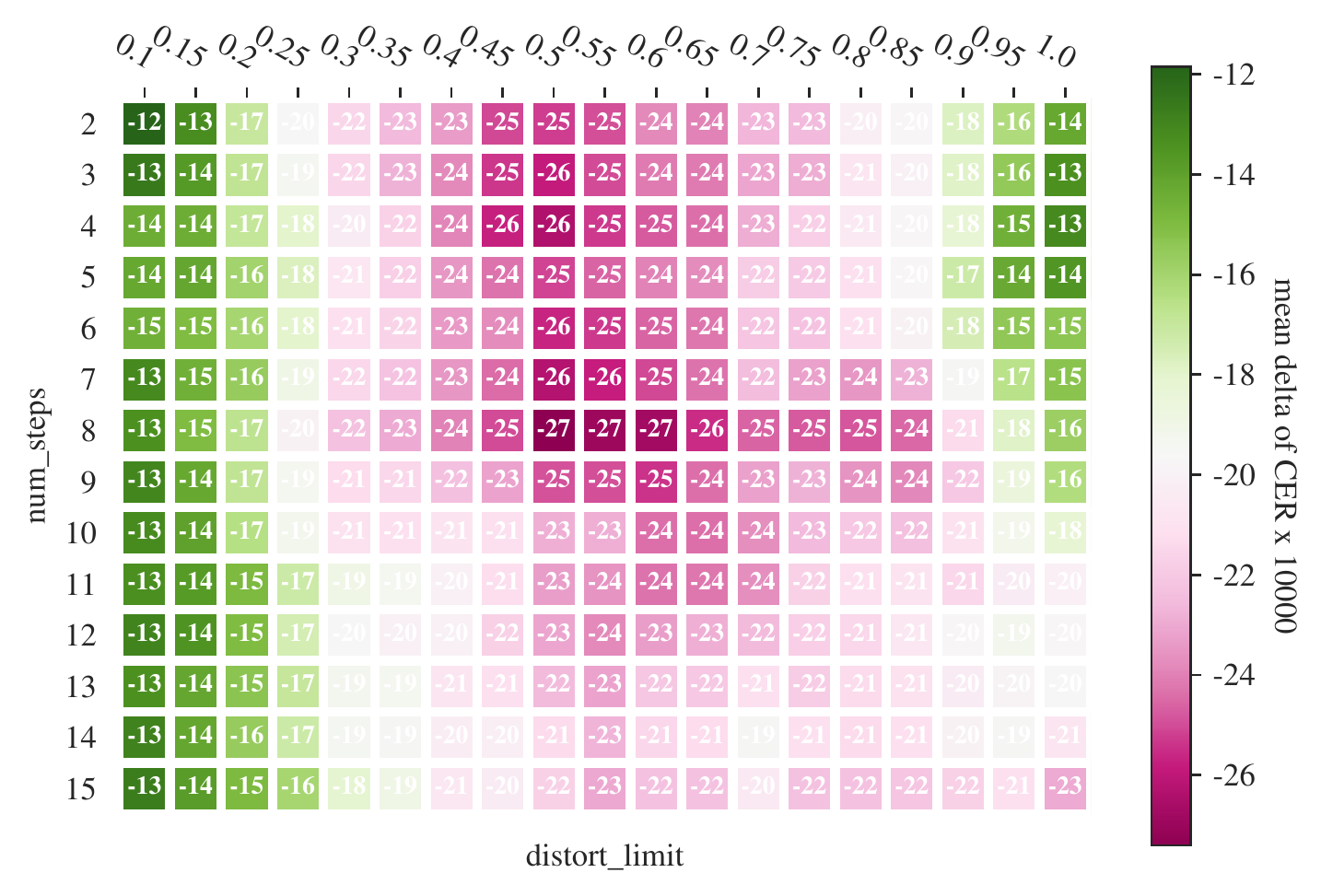}
    \caption{Mean improvement of CER for low-resolution grid distortion using a modified version of albumentations \cite{buslaev_AlbumentationsFastFlexible_2020}. Neighbouring cells are included in a cell's mean values to increase sample size. A CER delta of $1\%$ is displayed as 100.}
    \label{fig:distort-lowres}
\end{figure}

We also investigated low-res grid resolutions using the \textproc{num\_steps} and \textproc{distort\_limit} parameters of a version of albumentation's \textproc{GridDistortion} \cite{buslaev_AlbumentationsFastFlexible_2020} that we modified to normalize grid distortions (see Figure \ref{fig:distort-lowres}). \textproc{num\_steps} describes the number of divisions the distortion grid gets on each axis. \textproc{distort\_limit} models the maximum amount of distortion one grid vertex can receive. Due to the normalization of distortion amounts, no grid vertex will lie outside the image after distortion. Therefore, \textproc{distort\_limit} actually models a distortion variance.

Generally a \textproc{distort\_limit} of about $0.5$ works best, and we see an optimal CER percentage improvement of $0.27$ when using $8$ grid division steps and a decrease when using a higher number of steps.

\subsubsection{Image Degradation and Cutout}

\begin{figure}[ht!]
    \centering
    \includegraphics[scale=0.8]{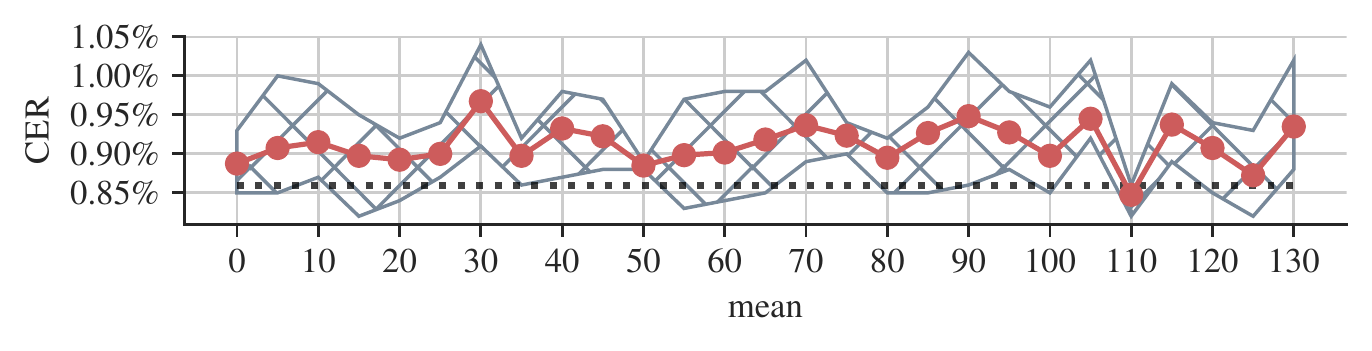}
    \caption{Influence on CER when adding noise when varying $\mu$ and using random $\sigma^2$ between 10 and 50. Dotted line is best unaugmented performance.}
    \label{fig:noise-mean}
\end{figure}

\begin{figure}[ht!]
    \centering
    \includegraphics[scale=0.8]{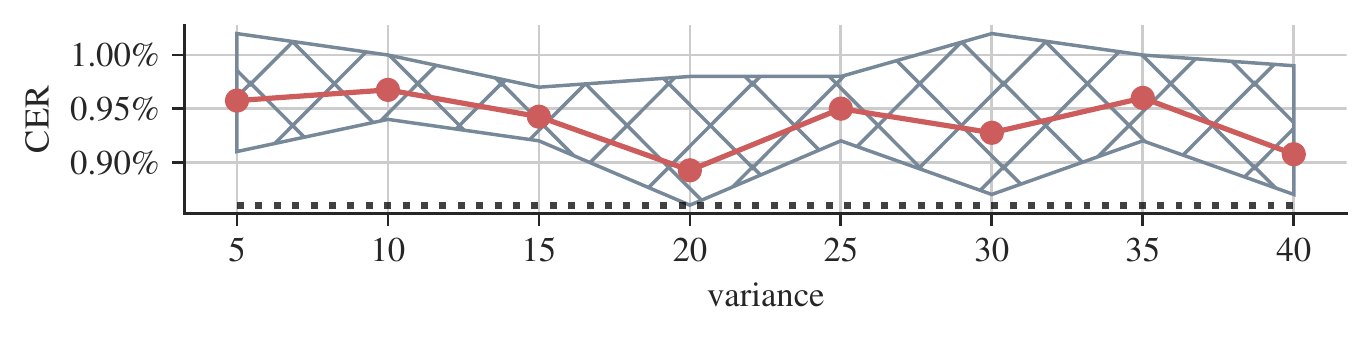}
    \caption{Influence on CER when adding noise with $\mu=15$ and varying $\sigma^2$. Dotted line is best unaugmented performance.}
    \label{fig:noise-variance}
\end{figure}

\paragraph{Gaussian Noise} As described in Section \ref{section:image-degradation}, we investigated the effect of adding Gaussian noise to the line images (which are encoded in gray values ranging from $0$ to $255$). Figures \ref{fig:noise-mean} and \ref{fig:noise-variance} show experiments with varying $\mu$ and $\sigma$ in terms of of the noise added. Good choices for $\sigma^2$ seem to be $20$ and $40$. One strikingly interesting mean value is at $\mu=110$. The mean improvement there is considerable ($0.15$, $n=4$), but it is highly inconsistent with neighboring values which makes it look unstable. We have not investigated this effect further.

\begin{figure}[ht!]
    \centering
    \includegraphics[scale=0.75]{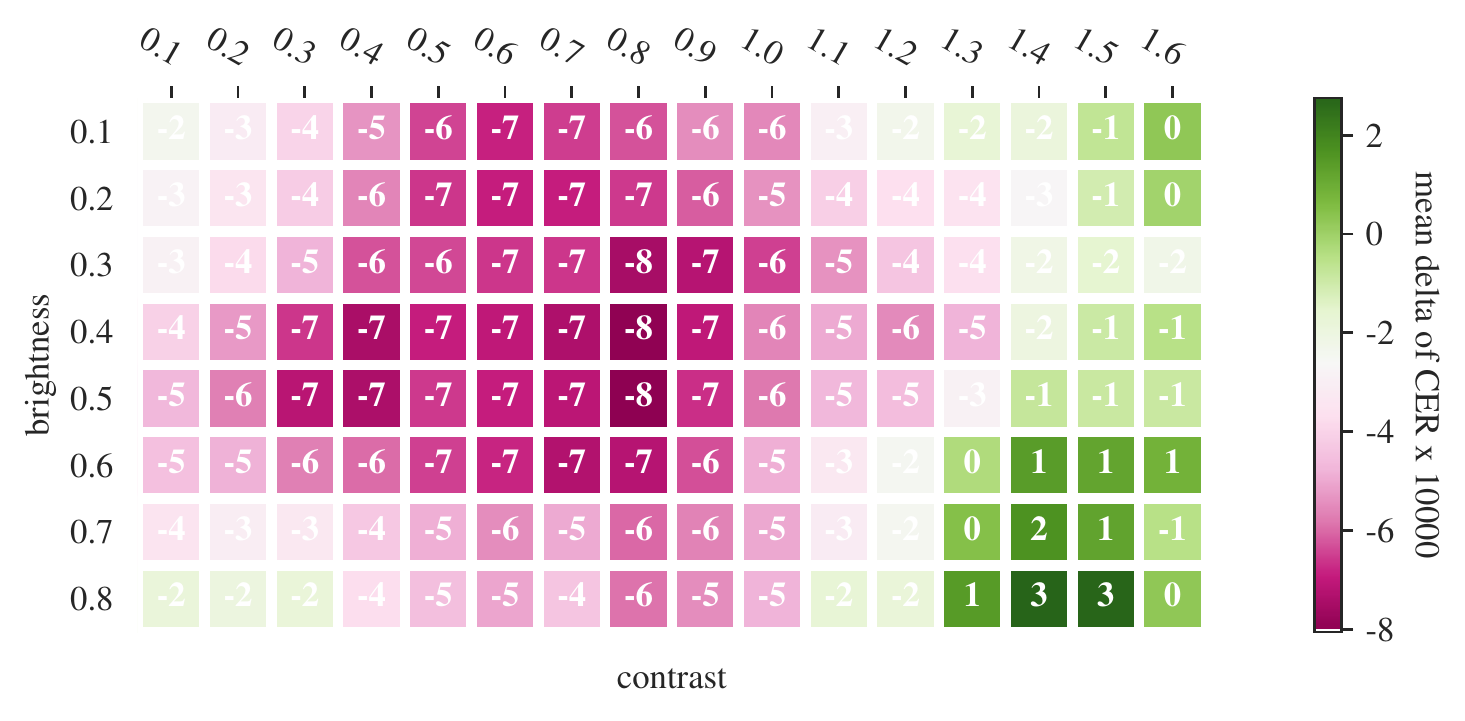}
    \caption{Mean improvement of CER when changing contrast and brightness of line images. Neighbouring cells are included in a cell's mean values to increase sample size. A CER delta of $1\%$ is displayed as 100.}
    \label{fig:contrast-and-brightness}
\end{figure}

\paragraph{Contrast and Brightness} Results of our experiments with contrast and brightness are shown in Figure \ref{fig:contrast-and-brightness}. As shown in Figure \ref{fig:contrast-and-brightness}, the area of parameters that give good performance is roughly circular around a brightness limit of $0.4$ and a contrast limit of $0.7$. Any combination near that pair of values seems to work well.

\begin{figure}[ht!]
    \centering
    \includegraphics[scale=0.55]{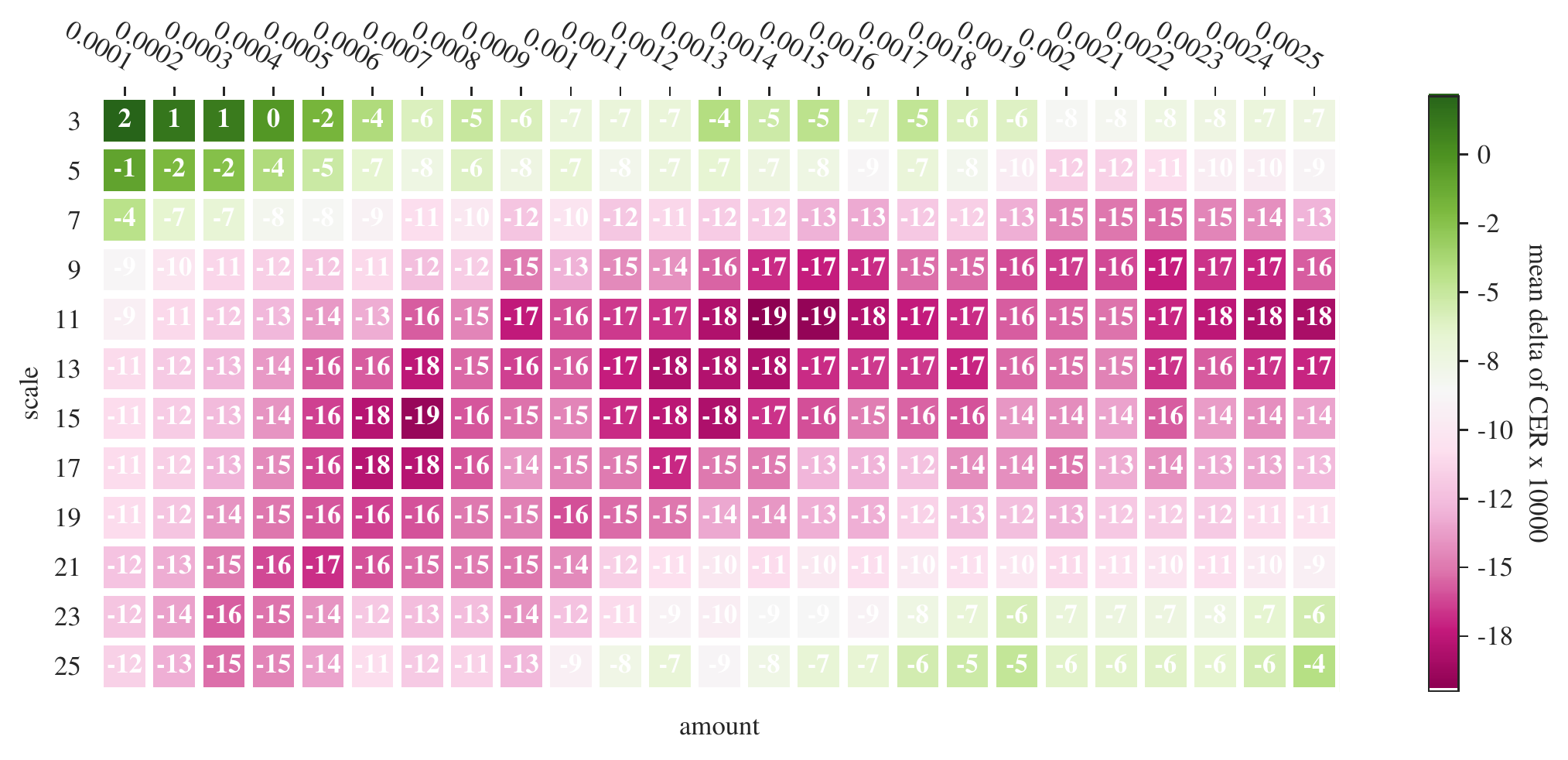}
    \caption{Mean improvement of CER for adding blotches using ocrodeg. Neighbouring cells are included in a cell's mean values to increase sample size. A CER delta of $1\%$ is displayed as 100.}
    \label{fig:blobs}
\end{figure}

\paragraph{Adding Blotches} As described in Section \ref{section:image-degradation}, we looked at the impact of adding random blotches %
by varying the two parameters $scale$ and $amount$ (see Figure \ref{fig:blobs}). Good values tend to lie at $9 <= scale <=15$. Two promising areas are at $amount=0.0007$ and the double of that amount, $amount=0.0014$.

\subsubsection{Amount of Augmentation}

One general question with data augmentation is how much augmentation should happen. In our case, the $min$ CER does not improve when generating more than twice our ground truth of 2807 lines (see Table \ref{tab:augmentation-1}). These results probably depend on the ground truth size as well as whether one applies a \emph{single} augmentation step or multiple such steps.

\begin{table}[htb!]
\center
\begin{tabular}{c c c c c} 
augmentation & $\mu$ CER & $\sigma$ CER & $min$ CER & $n$ \\
\hline
+50\% & 0.76\% & 0.07 & 0.67\% & 8 \\
+100\% & 0.74\% & 0.06 & \textbf{0.58\%} & 12 \\
+150\% & 0.69\% & 0.02 & 0.64\% & 8 \\
+200\% & 0.68\% & 0.05 & 0.62\% & 7 \\
+250\% & 0.68\% & 0.03 & 0.62\% & 8 \\
+300\% & \textbf{0.67\%} & 0.04 & 0.62\% & 8 \\
\end{tabular}
\caption{\label{tab:augmentation-1}CER when applying low-resolution grid with various amounts of augmentation.}
\end{table}

\subsection{Ablation Study}
\label{section:ablation}

Ablation studies, a term made popular through a tweet by Fran\c{c}ois Chollet in 2018\footnote{see \url{https://twitter.com/fchollet/status/1012721582148550662?s=20}.}, are a relatively recent development in machine learning, but are now considered by many an essential tool in understanding deep learning. Lipton and Steinhardt note: "Too frequently, authors propose many tweaks absent proper ablation studies, obscuring the
source of empirical gains. Sometimes just one of the changes is actually responsible for the improved
results" \cite{lipton_TroublingTrendsMachine_2019}.  They add that "this practice misleads readers to believe that all of the proposed changes are necessary" \cite{lipton_TroublingTrendsMachine_2019}, when in fact single components are responsible for the gains.

We now perform an ablation study on an OCR training pipeline we designed from the best components we found in the first part of this study, namely a high quality CRNN architecture and three well-performing data augmentation operations. In this ablation study, we also include training parameters such as batch size and clipping norm.

\subsubsection{Introduction}
\label{section:ablation-intro}

We now set out to test the generalizability of our findings up to this point by using it to train OCR models with a pipeline that employs the best components (e.g. the model architecture $\psi$ and the way of doing data augmentation) we found in the first part of this study. We train and test these models on an independent data set to obtain reliable evaluation results.

As a first step, we build a representative German Fraktur corpus by randomly sampling lines and ground truth from gt4hist's large \emph{DTA19} Fraktur corpus \cite{springmann_GroundTruthTraining_2018} and from the Neue Zürcher Zeitung (NZZ) \cite{strobel_ImprovingOCRBlack_2019}. We add NZZ ground truth to \emph{DTA19}'s 39 books, then sample evenly from each of the 40 works. Our corpus thus contains data from over 100 years from 40 sources. We use 30K random lines from this corpus as high-quality test set to reliably compute CERs of trained models. For training, we will use between 10K and 20K lines that are not in the test set.

\subsubsection{Typical CERs on the test set}

\begin{table}[!ht]
\centering
\begin{threeparttable}
\begin{tabular}{ l | l l l } 
& & & \\
name & engine & model & year\\
\hline
tess-frk & Tesseract 4.1.1 & frk & 2017 \\
tess-deu\-frak& Tesseract 4.1.1 & deu\_frak & 2018\\
tess-mh & Tesseract 4.0.0 & Fraktur, UB Mannheim\tnote{1} & 2019 \\
cal-ocrd & Calamari 0.3.5 & OCR-D calamari gt4hist\tnote{2} & 2020\\
cal-frk-2019 & Calamari 1.0.4 & calamari fraktur\_19th\_century\tnote{3} & 2019 \\
cal-frk-2020 & Calamari 1.0.4 & calamari fraktur\_19th\_century\tnote{3} & 2020\\
\end{tabular}
\begin{tablenotes}
\item[1] trained by Stefan Weil, see \url{https://ocr-d.de/en/models.html}.
\item[2] see \url{https://qurator-data.de/calamari-models/GT4HistOCR} and \url{https://github.com/qurator-spk/train-calamari-gt4histocr}.
\item[3] see \url{https://github.com/Calamari-OCR/calamari_models}.
\end{tablenotes}
\end{threeparttable}
\caption{Various pretrained Fraktur models we tested against.}
\label{table:gt4hist-testset-pretrained-engines}
\end{table}

\begin{table}[!ht]
\centering
\begin{tabular}{ l c | c c | c c | c c |} 
& & \multicolumn{2}{|c|}{otsu} & \multicolumn{2}{|c|}{sauvola} & \multicolumn{2}{|c|}{grayscale} \\
model & F & raw & hrm & raw & hrm & raw & hrm \\
\hline
tess-frk & & $7.99\%$ & $7.50\%$ & $7.83\%$ & $7.41\%$ & $6.49\%$ & $6.04\%$\\
tess-deu\-frak & & $11.61\%$ & $9.43\%$ & $10.94\%$ & $8.74\%$ & $11.57\%$ & $9.38\%$ \\
tess-mh & & $1.42\%$ & $1.28\%$ & $1.44\%$ & $1.31\%$ & $1.02\%$ & \textbf{0.88\%} \\

\hline
cal-ocrd & 1 & $1.33\%$ & $1.09\%$ & $1.43\%$ & $1.19\%$ & $2.41\%$ & $2.25\%$ \\
cal-ocrd & 5 & $1.20\%$ & $0.96\%$ & $0.66\%$ & \textbf{0.62\%} &  $2.48\%$ & $2.32\%$ \\

\hline
cal-frk-2019 & 1 & $1.83\%$ & $1.61\%$ & $2.36\%$ & $2.14\%$ & $2.89\%$ & $2.73\%$ \\
cal-frk-2019 & 5 & $1.46\%$ & \textbf{1.24\%} & $2.13\%$ & $1.91\%$ & $3.29\%$ & $3.13\%$ \\

\hline
cal-frk-2020 & 1 & $1.00\%$ & $0.89\%$ & $1.27\%$ & $1.13\%$ & $3.12\%$ & $3.06\%$ \\
cal-frk-2020 & 5 & $0.95\%$ & \textbf{0.79\%} & $1.20\%$ &  $1.03\%$ & $3.30\%$ & $3.25\%$ \\

\end{tabular}
\caption{CERs of pretrained Fraktur models on the 20K line \emph{DTA19}/NZZ test corpus, using binarized or grayscale input, with CER measured on unharmonized (raw) and harmonized transcriptions. The \emph{F} column indicates the number of folds used during prediction.}
\label{table:gt4hist-testset-pretrained}
\end{table}

\begin{figure}
    \centering
    \subfigure[]{\includegraphics[width=10cm]{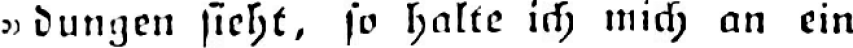}} 
    \subfigure[]{\includegraphics[width=10cm]{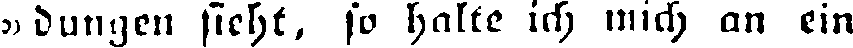}} 
    \subfigure[]{\includegraphics[width=10cm]{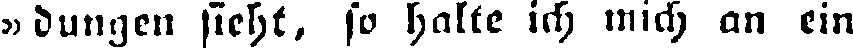}} 
    \caption{Line 184 from wackenroder\_herzensergiessungen from \emph{DTA19} which is part our test corpus, scaled to 48 px, and then (a) left as grayscale image, (b) binarized with an Otsu threshold, (c) binarized with a Sauvola threshold using a window size of 23. The ground truth transcription for this line is ``»dungen \longs ieht, \longs o halte ich mich an ein''.}
    \label{fig:ablation-bin-line}
\end{figure}

Tables \ref{table:gt4hist-testset-pretrained-engines} and \ref{table:gt4hist-testset-pretrained} are intended to give an indication of the level of difficulty involved in predicting this test set.

We evaluated two versions of Calamari's official fraktur\_19th\_century model (the older 2019 and the newer 2020 versions), as well as OCR-D's pretrained gt4hist model for Calamari. Note that at least two of these models used the complete \emph{DTA} Fraktur with over 240,000 lines in their training sets, which includes our full test set. These models are 5-fold ensemble models by default. We evaluated using only the first fold in order to get a measure of their performance without voting (see the column titled \emph{F}).

The column \emph{hrm} refers to the \emph{harmonization} (sometimes also called regularization) of transcription rules. This means that we remove obvious inconsistencies between the predicted text and the ground truth\footnote{A similar process is called \emph{harmonized transcriptions} in \url{https://github.com/tesseract-ocr/tesstrain/wiki/GT4HistOCR} and we adapt this term here.}. For example, some models transcribe \longs\enspace(long s) as such, others transcribe it as \emph{s} (e.g. the NZZ corpus) - we use only short s. We also regularized some quotation marks, various unicode codes for similar looking dashes, some characters can be hardly differentiated visually, duplicate whitespaces and various other sources of inconsistency\footnote{A more general discussion of these problems can be found in \cite{smith_ResearchAgendaHistorical_2018}}. The harmonized CER is lower and a better measure of the model's real accuracy than the raw (unharmonized) CER.  We will apply the same harmonizations on the training, validation and test sets from \emph{DTA19} when we train and evaluate our own model later.

Our harmonization produces transcriptions that are based on the following 144 character codec:

\begin{itemize}[noitemsep]
\item special characters: \texttt{\textvisiblespace\%\&*+=\_§$\dagger$}
\item punctuation: \texttt{!,-.:;?}
\item quotation marks: \texttt{\textquotestraightbase\textquotesingle«»}
\item brackets: \texttt{()[]<>}
\item slashes: \texttt{/|}
\item digits: \texttt{0123456789}
\item upper case letters: \texttt{ABCDEFGHIJKLMNOPQRSTUVWXYZ}
\item lower case letters: \texttt{abcdefghijklmnopqrstuvwxyzßø}
\item lower case letters with diacritica: \texttt{àáâãäåçèéêëîñòóôõöùûüĩũů\~{e}}
\item upper case letters with diacritica: \texttt{ÄÅÇÊÒÖØÙÚÛÜ}
\item ligatures: \texttt{Ææœ}
\item currencies: \texttt{€£}
\item greek: \texttt{$\alpha \acute{\alpha} \varepsilon I \kappa \mu \chi$}
\item r rotunda %
\item eth: \texttt{$\eth$}
\item scansion: \textipa{\u=\,} \textipa{\u\,}
\end{itemize}

Note that the absolute CERs in Table \ref{table:gt4hist-testset-pretrained} should be taken with some caution, and indeed they somewhat differ from previous similar studies \cite{reul_StateArtOptical_2019}. We have two hypotheses for this. First, our codec and harmonization rules differ from other studies. For example our codec contains 139 characters, which is considerably larger - and therefore more difficult - than the 93 character codec used by Reul et al. \cite{reul_StateArtOptical_2019}. Second, we scale all pre-binarized lines to a 48 px grayscale image and then re-binarize. This process seems to negatively impact the performance of many models. An alternative would be rescaling using nearest neighbors without rebinarizing. This might also be an indication that binarization based models' accuracies are somewhat brittle with regard to slight changes in the binarization process. Figure \ref{fig:ablation-bin-line} shows the input we give to the models for one exemplary line. 

We were surprised to find that Tesseract's models actually work much better on grayscale input than on binarized input. We were even more surprised by the relationships of binarization algorithm and accuracy for the Calamari models.. It seems that models are trained to a specific binarization context like Otsu (see cala-frk-2020) or Sauvola (see cala-ocrd) and break down when used with a different algorithm. Given that it is not common to document any preferred binarization methods for pretrained models, this seems like a very brittle and disturbing situation.

\subsubsection{Our model}

For training our own model, we use the same 139-characters codec described above.

In accordance with our findings in Section \ref{section:non-binarized-works-best}, and even though the original data from \emph{g4hist} is binarized, we do not re-binarize line images for the training of our own model. After rescaling them to the target line height of 48 px, we keep them as grayscale images\footnote{Note that even better CERs might be obtained if we had had test data that was not binarized before scaling to the line height at all.}. As a result, our models will work exceedingly bad (CER $> 10\%$) %
on binarized input, but very well on grayscale input.

\subsubsection{Stages}

\begin{table}[!htbp]
\centering
\begin{tabular}{ l l l l c } 
stage & name & description \\
\hline
1 & Calamari Default & Calamari default settings.\\
2 & Harmonize Transcriptions & More consistent ground truth \\
3 & NAS Architecture & Optimal network architecture \\
4 & Batch Size & Set batch size to 8 \\
5 & Clipping Norm & Set clipping norm to 0.001 \\
6 & DA Distort & Distortion data augmentation \\
7 & DA Blotches & Cutout data augmentation \\
8 & DA Contrast & Contrast data augmentation \\
9 & Ensemble Voting & Calamari 5-fold ensemble voting \\
10 & Remove NAS Architecture & Remove optimal network architecture \\
11 & Remove DA & Remove all data augmentation\\
\end{tabular}
\caption{Training stages.}
\label{table:gt4hist-ablation-stages}
\end{table}

Our model will be trained with a pipeline that we break apart into the 11 stages shown in Table \ref{table:gt4hist-ablation-stages} in order to form an ablation study:  stage 1 is a Calamari standard training without any special adjustments. Higher stages incrementally apply various changes, such as harmonization of transcriptions (stage 2), using the network architecture from the NAS from part 1 of this paper (stage 3), and so on.
The specific parameters used for stages 3, 6, 7, and 8 were obtained in part 1 of this paper. Table \ref{tab:psi-final} summarizes these settings (we use dropout of $0.5$ and, starting with stage 6, an augmentation of $+200\%$).

Note that the order of the stages is relevant and has been detected via various side experiments. For example, flipping the order of the data augmentation steps seems to worsen accuracy considerably.

\begin{table}[htb!]
\center
\begin{tabular}{l l} 
stage & parameters\\
\hline
3 & architecture N=124, R=1.5, K=6, P=1, M=\{650\} \\
6 & high resolution grid distortion with num\_steps=$8$, distort\_limit=$0.5$ \\
7 & ocrodeg blotches with fgblobs=$0.0009$ and fgscale=$9$ \\
8 & brightness\_limit=$0.2$, contrast\_limit=$0.9$ \\
\end{tabular}
\caption{\label{tab:psi-final}Specific parameters used for stages 3, 6, 7 and 8.}
\end{table}

Stages 6 and higher augment $n$ training lines to a total of $3n$ training lines. Good values for stages 3 and 4 have been found with smaller side experiments\footnote{For example, batch size 4 and 16 performed worse than batch size 8.}. Stages 10 and 11 both keep ensemble voting but remove network architecture and data augmentation, in order to investigate the negative impact this has for ensemble voting.

A note on sampling: for stages 1 up to 8 we estimate $\sigma$ to be below $0.04$. By training at least 16 models per stage, we therefore estimate the error in our evaluated CERs to be below $0.02\%$ with $95\%$ confidence. For stages 9 and higher, we estimate $\sigma$ at $0.01$. By training at least 4 ensemble models, we estimate our CER error at $0.01\%$ with $95\%$ confidence. We believe our study is more statistically solid than previous evaluations.

\begin{figure}[ht!]
    \centering 
    \includegraphics[scale=0.9]{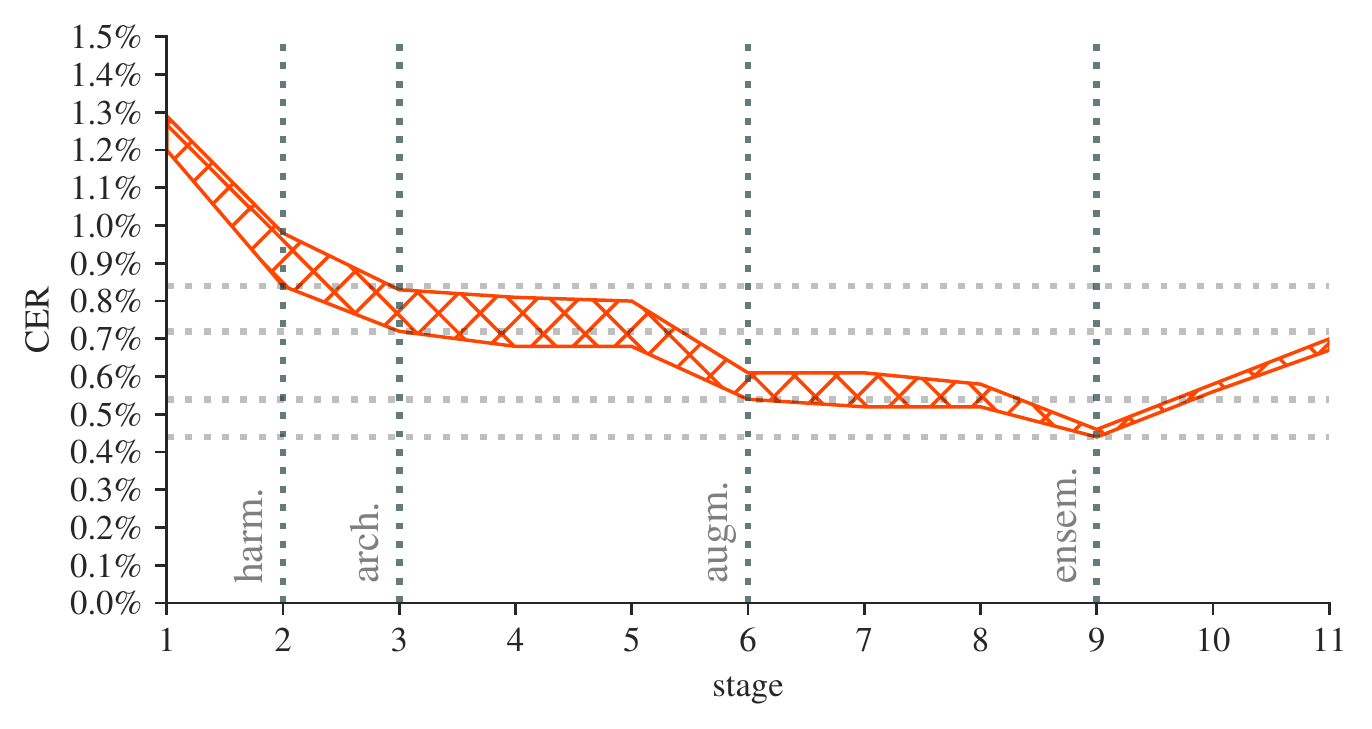}

	\caption{CERs for various stages. Plot of results of Table \ref{table:gt4hist-ablation-results}.}
	\label{fig:gt4hist-ablation-results}
\end{figure}

Table \ref{table:gt4hist-ablation-results} and Figure \ref{fig:gt4hist-ablation-results} show our results. In order to avoid confusion with relative percentages in the following tables and paragraphs, we introduce the acronym CERp, by which we denote the CER given as percentage. Thus, a CERp value of 5 corresponds to a CER of 5\%. 

We now investigate the improvements. After harmonization, our model's CERp of $0.91$ is similar to the $0.89$ CERp of the unvoted cala-frk-2020 otsu model. Since both models use no voting and the same Calamari default architecture, these numbers seem reasonable and demonstrate the baseline performance of training Calamari without further tricks. In our ablation study, we now observe four sources of improvement over this baseline. Together they reduce the mean CERp from $0.91$ to $0.45$. In detail:

\begin{enumerate}
\item The optimized network architecture yields an CERp improvement of $0.12$ (stages 3), even without any additional changes.
\item Optimized batch size and clipping norm improve the CERp by $0.06$ (stages 4 and 5).
\item Fine-tuned data augmentation yields a big CERp improvement of $0.17$ (stages 6 + 7 + 8). It is surprising to find that basically all benefit happens in the first DA step (distortion), whereas the following two steps hardly contribute. We experimented with various orders of applying DA, and always saw this behavior of the first DA stage. We also observed that some orderings and DA combinations that showed improvements when applied individually, worsened the results when applied in combination, which implies DA should be used very carefully and never in a "more is better" fashion. Also, the overall CER improvements on this corpus are still rather small compared to the values found in Section \ref{section:results-of-data-augmentation}.
\item Ensemble voting with 5 folds yields an CERp improvement of $0.1$. This is in accordance with the measurements on pretrained models in Table \ref{table:gt4hist-testset-pretrained}.
\end{enumerate}
Some changes, such as introducing clipping norm in stage 5, and contrast in stage 8, do not improve the best CERs. However, they seem to reduce the mean CERs and variance, thereby improving the probability of obtaining a better model with less trials. Therefore, they should be considered useful adjustments in a training pipeline.
In summary, the custom network architecture proves about as effective in reducing the CER as using ensemble voting with 5 folds. Data augmentation proves nearly twice as effective as ensemble voting, whereas batch size and clip size together are about half as effective.

\begin{table}[ht!]
\centering
\begin{tabular}{ l l c c c c c } 
stage & change & $\mu$ CERp & $\Delta \mu$ CERp & $\sigma$ CERp & $min$ CERp & n  \\
 \hline

1 &  & 1.25 &  & 0.022 & 1.2 & 20 \\
2 & + Harmonization & 0.91 & -0.35 & 0.037 & 0.84 & 20 \\
3 & + NAS Model & 0.78 & -0.12 & 0.029 & 0.72 & 20 \\
\\
4 & + Batch Size & 0.73 & -0.05 & 0.035 & 0.68 & 20 \\
5 & + Clipping Norm & 0.72 & -0.01 & 0.029 & 0.68 & 19 \\
6 & + DA Distort & 0.57 & -0.15 & 0.022 & 0.54 & 28 \\
7 & + DA Blotches & 0.56 & -0.01 & 0.019 & 0.52 & 32 \\
8 & + DA Contrast & 0.55 & -0.01 & 0.018 & 0.52 & 30 \\
9 & + Ensemble Voting & 0.45 & -0.10 & 0.006 & 0.44 & 7 \\
\\
10 & - NAS Model & 0.57 & 0.12 & 0.007 & 0.56 & 5 \\
11 & - DA & 0.68 & 0.11 & 0.011 & 0.67 & 6 \\
\end{tabular}
\caption{CERs (in percent) of various training stages. The column \emph{n} indicates the number of trained models.}
\label{table:gt4hist-ablation-results}
\end{table}

\subsubsection{Partial Data}

\begin{figure}[ht!]
    \centering 
    \includegraphics[scale=0.9]{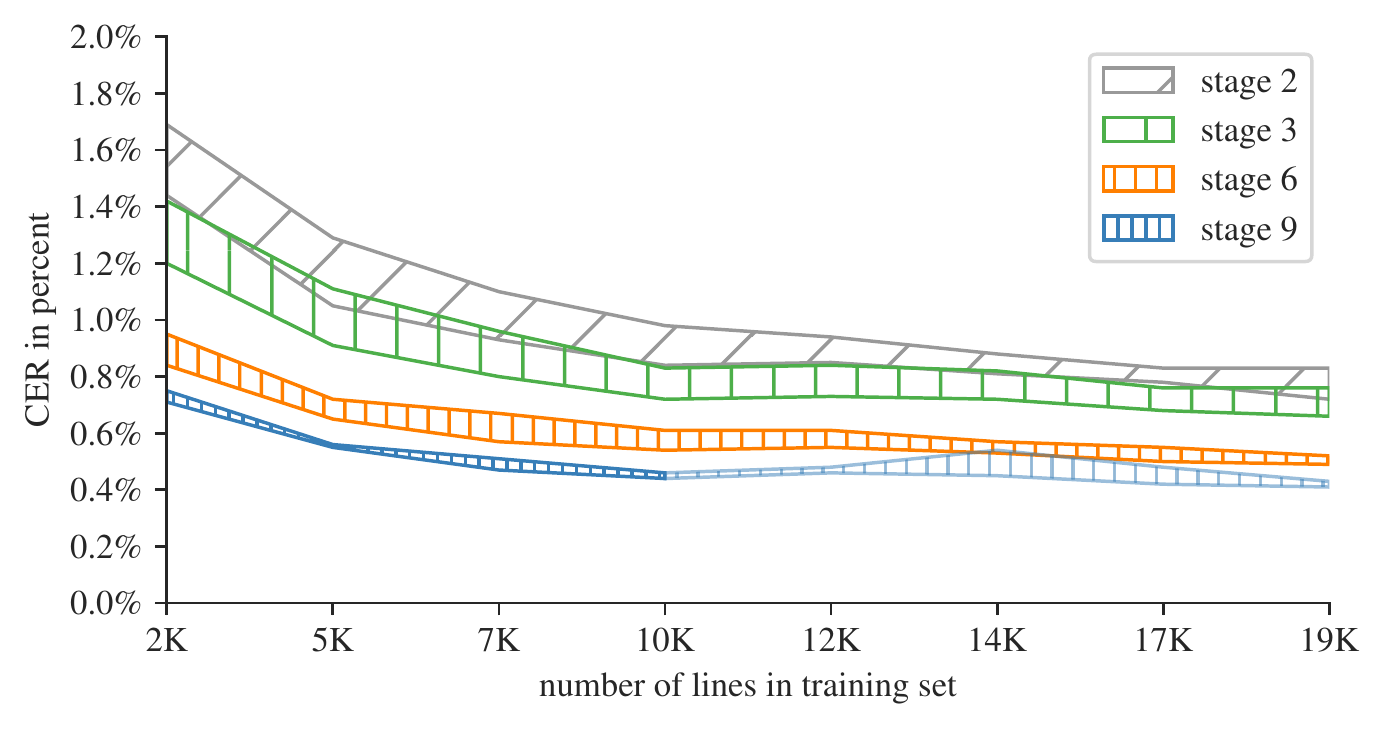}

	\caption{CERs for various combinations of stages and amount of training data. Plot of Table \ref{table:gt4hist-ablation-results-partial}. Models for stage9 and above 10K lines have been terminated after 100 hours of training (lighter blue area).}
	\label{figure:gt4hist-ablation-results-partial}
\end{figure}

In this section, we investigate the influence of the amount of training data on the achieved CERs. Table \ref{table:gt4hist-ablation-results-partial} and Figure \ref{figure:gt4hist-ablation-results-partial} show our results.
It is notable that while there may not always be an improvement in the obtained CERs with more training data, there is often a reduction in variance. For example, in stage 3, going from 10K to 14K, the $min$ CER hovers at $0.72\%$, but $\sigma$ drops from $0.029$ to $0.025$.

\begin{table}[ht!]
\centering
\begin{tabular}{ c c c c c c } 
 & lines of & &  & \\
stage & training data & $\mu$ CERp & $\sigma$ CERp & $min$ CERp & n \\
 \hline

2 & 2K & 1.55 & 0.081 & 1.44 & 16 \\
2 & 5K & 1.13 & 0.049 & 1.05 & 22 \\
2 & 7K & 0.99 & 0.047 & 0.93 & 17 \\
2 & 10K & 0.91 & 0.037 & 0.84 & 20 \\
2 & 12K & 0.88 & 0.022 & 0.85 & 19 \\
2 & 14K & 0.84 & 0.021 & 0.81 & 15 \\
2 & 17K & 0.81 & 0.020 & 0.78 & 8 \\
2 & 19K & 0.78 & 0.023 & 0.72 & 21 \\
\\
3 & 2K & 1.30 & 0.064 & 1.20 & 16 \\
3 & 5K & 0.99 & 0.058 & 0.91 & 16 \\
3 & 7K & 0.87 & 0.046 & 0.80 & 16 \\
3 & 10K & 0.78 & 0.029 & 0.72 & 20 \\
3 & 12K & 0.79 & 0.027 & 0.73 & 26 \\
3 & 14K & 0.75 & 0.025 & 0.72 & 16 \\
3 & 17K & 0.72 & 0.022 & 0.68 & 17 \\
3 & 19K & 0.71 & 0.026 & 0.66 & 22 \\
\\
6 & 2K & 0.91 & 0.027 & 0.84 & 16 \\
6 & 5K & 0.68 & 0.019 & 0.65 & 19 \\
6 & 7K & 0.61 & 0.024 & 0.57 & 16 \\
6 & 10K & 0.57 & 0.022 & 0.54 & 28 \\
6 & 12K & 0.58 & 0.014 & 0.55 & 22 \\
6 & 14K & 0.55 & 0.012 & 0.53 & 18 \\
6 & 17K & 0.52 & 0.015 & 0.50 & 15 \\
6 & 19K & 0.50 & 0.008 & 0.49 & 10 \\
\\
9 & 2K & 0.73 & 0.013 & 0.71 & 8 \\
9 & 5K & 0.55 & 0.005 & 0.55 & 7 \\
9 & 7K & 0.49 & 0.013 & 0.47 & 8 \\
9 & 10K & 0.45 & 0.006 & 0.44 & 7 \\
9 & 12K & 0.47 & 0.008 & 0.46 & 7 \\
9 & 14K & 0.47 & 0.033 & 0.45 & 6 \\
9 & 17K & 0.44 & 0.022 & 0.42 & 6 \\
9 & 19K & 0.42 & 0.006 & 0.41 & 6 \\

\end{tabular}
\caption{Relation of training data amount and CER (given in percent). Models for stage9 and above 10K lines have been terminated after 100 hours of training. Column \emph{n} indicates the number of trained models.}
\label{table:gt4hist-ablation-results-partial}
\end{table}

As can be seen in stages 3, 6, and 9, the benefits of good network architecture and data augmentation cannot be easily compensated for with more training data. For example, the average model from stage 3 with 19K lines of ground truth (CER $0.71\%$) shows roughly the same performance as the average model from stage 9 using only 2K of ground truth (CER $0.73\%$), i.e. a similar performance with only about a tenth the amount of ground truth.

\subsubsection{Reassessing Earlier Decisions}
\label{section:stage8-fine-tuning}

Now that we have a robust model incorporating all necessary training steps in stage 8, we can validate some central design decisions and try to find further improvements.
We first check the amount of augmentation (see Table \ref{table:stage8-augmentation}). As it turns out, adding 6 times more data instead of 2 times (i.e. generating about 60K lines to the given 10K lines) considerably improves the model again and brings the best models (\emph{min} CER) to $0.46\%$, which is very near the 10K stage 9 models. Increasing the rate from +600\% further to +900\% seems to have little consequence in terms of CERs though.

\begin{table}[ht!]
\centering
\begin{tabular}{ l c c c  } 
augmentation & $\mu$ CERp & $min$ CERp & n \\
 \hline
+200\% & 0.55 & 0.52 & 30 \\
+300\% & 0.54 & 0.50 & 22 \\
+400\% & 0.51 & 0.49 & 12 \\
+500\% & 0.52 & 0.49 & 14 \\
+600\% & 0.50 & 0.46 & 15 \\
+900\% & 0.49 & 0.45 & 12 \\
\end{tabular}
\caption{Effects of using more data augmentation in stage 8. +200\% is the default used in Table \ref{table:gt4hist-ablation-results}. The column \emph{n} indicates the number of trained models.}
\label{table:stage8-augmentation}
\end{table}

Finally we look at varying the depth of the network architecture $K$ to validate our initial choice of $K=6$. As discussed in Section \ref{section:results-nas}, $K=12$ specifically might work even better than $K=6$. Table \ref{table:stage8-k} shows that this does not seem to be the case with this larger corpus. In fact, we now see that $K=6$ is exactly the point where no further improvements happen.
\begin{table}[ht!]
\centering
\begin{tabular}{ l c c c c  } 
$K$ & $\mu$ CERp & $\sigma$ CERp & $min$ CERp & n \\
 \hline
2 & 0.59 & 0.01 & 0.56 & 18 \\
4 & 0.56 & 0.01 & 0.54 & 18 \\
6 & 0.55 & 0.02 & 0.52 & 30 \\
8 & 0.55 & 0.02 & 0.52 & 17 \\
10 & 0.54 & 0.02 & 0.51 & 18 \\
12 & 0.55 & 0.02 & 0.52 & 17 \\ 
\end{tabular}
\caption{Effects of varying $K$ in stage 8. The column \emph{n} indicates the number of trained models.}
\label{table:stage8-k}
\end{table}
We also investigated whether we might gain better CERs when combining high augmentation \emph{and} deeper networks, but found that this is not the case (\ref{table:stage8-k-aug}). In fact, the results seem to get worse with higher $K$. %

\begin{table}[ht!]
\centering
\begin{tabular}{ l c c c c  } 
$K$ & $\mu$ CERp & $\sigma$ CERp & $min$ CERp & n \\
 \hline
8 & 0.47 & 0.01 & 0.46 & 4 \\
10 & 0.50 & 0.02 & 0.48 & 2 \\
\end{tabular}
\caption{Effects of varying $K$ in stage 8 when setting augmentation to +900\% at the same time.}
\label{table:stage8-k-aug}
\end{table}

\section{Conclusion}

Based on available research on current state of the art OCR and HTR networks and through extensive grid searches, we have investigated the optimal settings for line image formats, CRNN network architectures, elastic deformations, grid deformations, cutout augmentations and several other building blocks of a state of the art OCR training pipeline.
Building upon this set of components and parameters, we designed a full OCR training pipeline, which we then evaluated on an independent data set of 20K randomly selected lines from the big \emph{DTA19} Fraktur corpus. We provide a detailed ablation study to demonstrate the influence of each major component for the overall pipeline's accuracy. A number of these operations can easily be incorporated into existing training pipelines.
Given this pipeline, we have also demonstrated that about 10K lines of training data and the open-source framework Calamari are sufficient to get similar results to very recent state of the art closed-source solutions trained on four times the training data \cite{strobel_HowMuchData_2020}. The model we trained also outperformed various other pretrained state of the art models.
Finally we were able to show that it is possible to avoid the higher inference cost of ensemble models by trading more training time for prediction time. Using considerable data augmentation, we obtained a model that predicts much faster than ensemble models but still shows similar CERs (namely $0.46\%$ CER with 10K ground truth $+600\%$ data augmentation vs. $0.44\%$ CER with a 5-fold ensemble model and 10K ground truth $+200\%$ data augmentation). In this paper, we also aimed at providing informative evaluations by  consistently sampling from model distributions (see Section \ref{section:model-dist}) on the one hand and using ablation studies (see Section \ref{section:ablation})  on the other hand. Both practices are not at all common yet, but address two common shortcomings of existing DNN studies \cite{lipton_TroublingTrendsMachine_2019, berg_TrainingVariancePerformance_2017}.

Given the results from Section \ref{section:non-binarized-works-best}, we strongly encourage the OCR community to reevaluate the role of binarization in OCR pipelines and ground truth datasets. Our results clearly show that for a wide variety of model architectures -- and at least the two most commonly used line heights -- non-binarized input reliably outperforms a high-quality (Sauvola) binarized input for CRNN architectures when used on historical material with bad paper quality. The latter is a context where binarization should actually prove the most useful and show the most benefit, but actually seems to do the opposite. Even more disconcerting is the insight that our results seem to show that models trained on a specific binarization algorithm will yield rather bad accuracy when later used with a different, but possibly better, binarization algorithm. Therefore, binarized-only datasets seem -- at best -- like a missed opportunity to obtain even better neural network models, and -- at worst -- like in intrinsic bias in training architectures.

We hope that future studies into OCR architecture might leverage smarter Neural Architecture Search approaches -- e.g. using reinforcement learning or population-based training -- and incorporate newer and more complex DNN components, such as for example residual connections and more recent LSTM or GRU layer designs.

\section{Acknowledgment}

This research was supported by the Deutsche Forschungsgemeinschaft (DFG, German Research Foundation), project number BU 3502/1-1.

Computations in this paper were performed on the cluster of the Leipzig University Computing Centre.

\medskip

\printbibliography

\end{document}